\documentclass[preprint,12pt]{elsarticle}

\usepackage{amsthm}
\usepackage{graphicx}
\usepackage[colorinlistoftodos]{todonotes}
\usepackage[colorlinks=true, allcolors=blue]{hyperref}
\usepackage{comment}
\usepackage{subfig}
\usepackage{caption}
\usepackage{tikz}
\usepackage{amsmath, amssymb}
\usepackage{appendix}
\usepackage{natbib}
\usetikzlibrary{decorations.text}
\usetikzlibrary{shapes.geometric}
\usetikzlibrary{positioning}
\usepackage{amsthm}

\DeclareMathOperator{\SatAgg}{SatAgg}
\newtheorem{theorem}{Theorem}[section]
\newtheorem{lemma}[theorem]{Lemma}
\newtheorem{proposition}[theorem]{Proposition}
\newtheorem{cor}[theorem]{Corollary}

\newtheorem{definition}{Definition}[section]

\newtheorem{example}{Example}[section]

\DeclareMathOperator*{\argmin}{arg\,min}

\begin{document}

\begin{frontmatter}

\title{A Semantic Framework for Neuro-symbolic Computation}

\author{Simon Odense$^1$}
\ead{simon\_odense@sfu.ca}

\author{Artur d'Avila Garcez$^2$}
\ead{a.garcez@city.ac.uk}

\address{$^1$Simon Fraser University, Vancouver, Canada \\$^2$ City, University of London, UK}

\begin{abstract}
The field of neuro-symbolic AI aims to benefit from the combination of neural networks and symbolic systems. A cornerstone of the field is the translation or \textit{encoding} of symbolic knowledge into neural networks. Although many neuro-symbolic methods and approaches have been proposed, and with a large increase in recent years, no common definition of encoding exists that can enable a precise, theoretical comparison of neuro-symbolic methods. This paper addresses this problem by introducing a semantic framework for neuro-symbolic AI. We start by providing a formal definition of \textit{semantic encoding}, specifying the components and conditions under which a knowledge-base can be encoded correctly by a neural network. We then show that many neuro-symbolic approaches are accounted for by this definition. We provide a number of examples and correspondence proofs applying the proposed framework to the neural encoding of various forms of knowledge representation. Many, at first sight disparate, neuro-symbolic methods, are shown to fall within the proposed formalization. This is expected to provide guidance to future neuro-symbolic encodings by placing them in the broader context of semantic encodings of entire families of existing neuro-symbolic systems. The paper hopes to help initiate a discussion around the provision of a theory for neuro-symbolic AI and a semantics for deep learning.
\end{abstract}

\end{frontmatter}


\sloppy

\section{Introduction}

\label{neuro-symbolic Integration}
 The separation of the AI community into two areas of research - connectionism versus symbolic AI - has resulted in two largely separate types of AI systems: deep neural networks and agent-based systems, respectively. Deep neural networks use massively parallel computational models known as artificial neural networks (ANNs). 
 ANNs use very efficient message passing to model statistical regularities in large amounts of data. This makes them very effective at making predictions about unseen data, solving practical problems such as facial recognition and audio prediction involving high-dimensional, multi-modal and weakly-correlated data streams. 
 By contrast, symbolic AI seeks to develop systems that reason explicitly about the world using a set of generally-applicable rules that serve to manipulate pre-defined symbols to reach a  conclusion. Parallels have been drawn between the connectionist and symbolic paradigms of AI and Daniel Kahneman’s research on human reasoning and decision making and so-called AI systems 1 and 2 \cite{Kahneman2011}. 
 
 Despite all the recent success of deep learning, it seems clear now that ANNs often struggle with learning and reasoning about abstract properties and concept hierarchies. A better approach should allow for what has been learned to be described in ways that humans can understand, to be efficiently re-used in a different but related situation, and to be reasoned about and safely improved upon, as argued in \cite{3rdwave}. Even with headline results such as the super-human capability of deep neural networks, e.g. at playing games, small changes made to the game environment can cause the network to underperform or fail altogether \cite{deepreinforcement}. In the case of large language models, it is known that certain changes made to the input prompt may cause the network to produce wrong outputs, which became known as hallucinations \cite{AIindex}. Such results suggest an inability of standard neural networks to learn the generally-applicable principles that could be applied robustly to different games or to extrapolate beyond the training data to learn the key abstract concept relations that may guard against hallucinations. 
 

 Specific neural network architectures are therefore designed to accomplish specific tasks related to specific properties of datasets, e.g. image translation invariance in the case of Convolutional Neural Networks (CNNs) \cite{conv_approx}. Thus, a major research question persists: how might a neural network be designed to learn and describe a general principle? Suppose that a dataset satisfies a set of logical propositions $L$. One would expect that a network that has learned to represent, or \emph{encode} $L$ should perform better on that dataset than a network that has not. But what does it mean for a network to encode a logical proposition? This is the question that we shall formalize in this paper. It is a pre-requisite to answering the earlier, bigger question, under the assumption that the general principle can be described by computational logic.  
 
 The question of encoding knowledge in neural networks pertains to the domain of neuro-symbolic AI \cite{NeSybook1,NeSyBook}. Countless techniques for combining neural networks and symbolic AI have been developed. However, as we shall see, there does not exist one unifying framework for neural encoding. Not only does this make the development of a theory for neuro-symbolic AI impossible, it also makes it difficult to compare the many different techniques in the literature as they may use different notions of encoding and logical equivalence. 

The field of neuro-symbolic AI has developed with the objective of combining the strengths of neural networks and symbolic systems, including the integration of learning and reasoning systems containing neural and symbolic components \cite{NeSybook1}. In neuro-symbolic AI, most of the attention has been focused on the development and practical evaluation of such systems that might benefit from data and knowledge represented in various forms, and the extraction of symbolic knowledge from trained neural networks, this latter area having made fundamental contributions to the field of explainable AI \cite{xaitutorial}. 
That is to say, methodologies have been developed generally independently with a focus on practical concerns and empirical evaluation. By contrast, this paper is focused on making a contribution towards the provision of a unifying underlying foundation for neuro-symbolic AI.
 
This paper introduces a framework for the encoding of the semantics of logic into neural networks. Through the formal definitions of \emph{semantic encoding} introduced in the paper, we emphasise the need for neuro-symbolic AI that is based on provably sound translations of knowledge into networks. 
The framework establishes correspondences between logical systems and classes of neural networks. It is expected to serve as a foundation for the future development of a theory for neuro-symbolic AI by defining a general and yet precise set-theoretic notation required for the development of such a theory. 

In a nutshell, this paper will offer a set of definitions of the components and conditions that must be satisfied for a knowledge-base to be encoded correctly by a neural network. The contribution of the paper is three-fold, offering:

\begin{itemize}

\item a unifying framework for semantic encoding upon which a theory of neuro-symbolic AI can be developed in the future; 

\item a tool for the systematic description of a large number of existing neuro-symbolic approaches w.r.t. a choice of encoding and aggregation;  

\item a formalization capable of providing a certain guidance to future neuro-symbolic encodings by placing them in the broader context of the semantic encoding of families of neuro-symbolic systems. 

\end{itemize}


While the differences in approach of a large number of loosely-coupled neuro-symbolic systems may make the development of a all-encompassing framework unlikely, this paper shows that many of the more tightly-coupled neuro-symbolic approaches can be formalized within the proposed framework. This is because, at a fundamental level, semantic encoding works by mapping the state of a neural network to semantic information. Despite the differences in network architecture, encoding technique, and the logic being encoded, our framework reveals that many neuro-symbolic approaches form a semantic encoding according to three basic components: a choice of mapping, aggregation and encoding function. In broad strokes, we define a semantic encoding of a knowledge-base into a neural network as a mapping such that each state of the network corresponds to  semantic information obtained from the knowledge-base. By aggregating the information contained in the states that a neural network converges to, we are left with the models of the knowledge-base being encoded by the network.\footnote{The use of the term \emph{models} here refers to the formal definition of a model in logic, that is, an assignment of truth-values (\emph{True} or \emph{False}) mapping a knowledge-base to \emph{True}.} This will be shown to encompass a large number of the techniques found in the literature, as well as to be general enough to apply to under-explored avenues of semantic encoding such as using neural networks with more complicated dynamic systems.

The remainder of the paper is organized as follows. In Section 2, we cover the necessary background and specify the notation needed to define semantic encoding used throughout the paper. 
In Section 3, we provide the formal definitions and examples of semantic encoding. In Section 4, we show that many neuro-symbolic approaches form a semantic encoding according to the unifying formal definitions. In Section 5, we conclude with a discussion of results and directions for future research. An Appendix contains the neuro-symbolic correspondence proofs obtained with the use of the proposed framework and the details of auxiliary results.

\section{Notation and Background}

We start with a simple motivating example before introducing the more technical concepts and required notation.

\begin{example}

 
Consider the simplest feed-forward neural network with a single hidden neuron, two input neurons ($A$ and $B$) and one output neuron ($C$), shown in Figure \ref{nnfig:1}. Let us assume for now that all neurons in the network take binary values $\{0,1\}$ and each neuron is a threshold perceptron (updated according to the output of a step function). Let each input and output neuron denote a propositional atom ($A,B,C$) with values in $\{0,1\}$ mapped to truth-values $False$ and $True$ (this localist binary representation to be extended to a distributed vector representation and many-valued atoms later). Each state of this network represents an assignment of truth-values to the atoms. With its current weights and biases in $\{-1,1\}$ as shown in Figure \ref{nnfig:1} (also to be extended to real numbers later), the network encodes the knowledge-base (or program) $P=\{C\leftarrow A; C\leftarrow B; A\leftarrow\}$ meaning that \textit{C} is $True$ if \textit{A} is $True$, \textit{C} is $True$ if \textit{B} is $True$, and \textit{A} is $True$ (in this case, we say that \textit{A} is a fact). The first two rules can be combined and described as: if A or B holds then C also does. Given, for example, a network state $(A,B,h,C)=(0,1,0,1)$, denoting that neuron A has activation value zero, neuron B has activation value 1 and so on, updating the state of the network in the usual way should produce state $(1,0,1,0)$. The bias of $A$ shown in Figure \ref{nnfig:1} as being equal to $1$ makes the new state of $A$ equal to $1$ given any input to $A$ in $\{0,1\}$. For the same reason, the state of $B$ becomes $0$ with the bias of $-1$. The previous state of $B=1$ makes $h=1$, and the previous state of $h=0$ makes $C=0$. Updating a second time gives the state $(1,0,1,1)$. This is because $A=1$ produces $h=1$, and the previous $h=1$ gives $C=1$. State $(1,0,1,1)$ then produces state $(1,0,1,1)$, that is a stable state equivalent to mapping $(A,B,C)$, starting from $(False,True,True)$, to $(True,False,True)$, which is the same result as calculating the least fixed-point of $P$ \cite{fixedpointlogic}. This correspondence between the network states given a set of weights and the fixed-point semantics of a logic program, once proven, makes the neural network semantically equivalent to the logic program. Implementing the least fixed-point operator $T_P$ of a logic program in a neural network is a common method for encoding symbolic knowledge into neural networks \cite{NeSybook1,CILP,original2}, first introduced in \cite{original1}.\footnote{Differently from the example shown here, in \cite{original1} an auto-associative recurrent network is used: input neurons (A and B) are repeated in the output layer of the network and output neuron C is repeated in the input layer. Although the representation used here is more compact, the basic idea is the same.} 
 
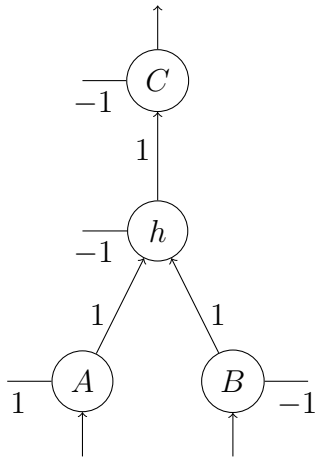
\begin{figure}[htp]

\centering
\begin{tikzpicture}

\useasboundingbox (0,0) rectangle (4,6);

 \node[shape=circle,draw=black, anchor=center,minimum size=0.8cm] (a) at (1,1){$A$};
  \node[shape=circle,draw=black, anchor=center,minimum size=0.8cm] (b) at (3,1){$B$};
 
  \node[shape=circle,draw=black, anchor=center,minimum size=0.8cm] (h) at (2,3){$h$};

  \node[shape=circle,draw=black, anchor=center, minimum size=0.8cm] (c) at (2,5){$C$} ;
  
  \coordinate (b_1) at (0,1);
  \coordinate (b_2) at (1,3);
  \coordinate (b_3) at (1,5);
  \coordinate (b_4) at (4,1);

\coordinate (i_1) at (1,0);
\coordinate (i_2) at (3,0);
\coordinate (o_1) at (2,6);

\draw[->] (a) -- (h) node[ above, xshift=-0.8cm, yshift=-1.39cm]{$1$};

\draw[->] (b) -- (h) node[ above, xshift=0.8cm,yshift=-1.39cm]{$1$};

\draw[->] (h) -- (c) node[ above, xshift=-0.2cm,yshift=-1.2cm]{$1$};

\draw[-] (b_1) -- (a) node[near start, below, xshift=0.0cm]{$1$}; 
\draw[-] (b_2) --  (h) node[near start, below,xshift=-0.0cm]{$-1$};
\draw[-] (b_3) --  (c) node[near start, below,xshift=-0.0cm]{$-1$};
  
\draw[-] (b_4) -- (b) node[near start, below,xshift=-0.0cm]{$-1$};

 \draw[->] (i_1) -- (a) node[near start, below, xshift=0.0cm]{}; 
 \draw[->] (i_2) -- (b) node[near start, below, xshift=0.0cm]{}; 
 \draw[->] (c) -- (o_1) node[near start, below, xshift=0.0cm]{}; 

\end{tikzpicture}
\caption{A simple feed-forward neural network encoding a knowledge-base containing rules \textit{C if A}, written  $C\leftarrow A$, \textit{C if B}, written $C\leftarrow B$, and fact \textit{A}, written $A\leftarrow $. The parameters of the network (weights and biases) are shown next to the arrows in the diagram. With bias $1$, neuron \textit{A} will always produce output $1$ (we say that \textit{A} is \emph{activated} in this case) for any input in $\{0,1\}$, given a step function as activation function. With bias $-1$, neuron \textit{B} will output $0$ for every input. Activating either \textit{A} or \textit{B} will always activate neuron \textit{h}, since the weight ($1$) from either \textit{A} or \textit{B} to \textit{h} is equal  to (or larger than) the negative of the bias of \textit{h}. Finally, activating \textit{h} also activates \textit{C}, for the same reason as above.}
\label{nnfig:1}
\end{figure}

Another notion of equivalence that is not dependent on the fixed-point semantics of logic programming - and therefore applicable to other logic representations - can be found in the encoding of propositional and non-monotonic logic into the energy function of symmetric neural networks \cite{penalty,SonTran}. Again, each state of the neural network represents a propositional truth-assignment and the network encodes a knowledge-base in the sense that minima of the network's energy function correspond to the models of a knowledge-base. In probabilistic approaches such as \cite{MLN,deepproblog}, the models of a knowledge-base are encoded into the states of a distribution with non-zero probability. Although these might seem to be radically different notions of encoding, they all share commonalities which we shall be able to formalize using the framework introduced in this paper. For one, each state of the neural network represents semantic information. Furthermore, the models of the knowledge-base correspond to stationary points of the neural network. For example, for Horn clauses, the model of a logic program is the stationary-point of fixed-point operator $T_P$, an energy-based network always converges to a minimum of its energy function \cite{hopfield}, and probabilistic approach Markov Logic Networks (MLNs) are guaranteed to converge to their stationary distribution \cite{MLN}. There are neuro-symbolic techniques that do not use semantic information in neural encodings, e.g. \cite{TPRorig}. However, semantic methods make up a large portion of neural encodings in the literature and, for this reason, will be the focus of this paper. 

%

\end{example}

\subsection{Artificial Neural Networks}
The central object of study in neural computation is the \textit{artificial neural network} (ANN), which we will refer to simply as a neural network. A neural network is a highly parallel model consisting of simple computational units called neurons which communicate via a set of weighted connections. Artificial neural networks were developed as a computational model meant to replicate the behaviour of the biological neural networks found in the nervous system \cite{mcculloch}. It is interesting to note that already in their original paper, McCulloch and Pitts' objective was to map such models to logical properties and operations. Although substantial differences exist between the neurons found in the nervous system and the idealized neurons used in artificial neural networks \cite{hodgkin}, the central idea is preserved: a neuron receives input from the neurons to which it is connected. If the total input is greater than a certain threshold the neuron will fire, sending the signal to all the other neurons that it is connected to. Hence, an ANN is a computational model. We can construct a graph from a neural network by adding nodes for each neuron and an edge between nodes if there is a connection between the corresponding neurons in the neural network. If the resulting graph contains no cycles then we say that the network is \textit{feed-forward}, and if it contains cycles we say that the network is \textit{recurrent}. Each neuron in a neural network is given a label $i\in \mathbb{N}$ and its activation value is denoted with a variable, $x_i$. In the basic conception of ANNs, the input to $x_i$ is a weighted sum of the values of the neurons connected to $i$, in other words the input to a neuron $i$ is $\sum\limits_j w_{ji}x_j + b_i$, where $w_{ji}\in \mathbb{R}$ is the connection weight from neuron $j$ to $i$, and $b_i\in \mathbb{R}$ is an additional parameter called the bias. The input to the neuron is then passed to the \textit{Heaviside step function}, defined by $f(x)=1$ if $x\geq 0$, and $f(x)=0$ otherwise. Given values for each neuron, we can use this function to calculate an updated value, $x_i'$, for neuron $i$ as $x_i'=1$ if $\sum\limits_j w_{ji}x_j + b_i\geq 0$, and $0$ otherwise. 

The basic neuron described above can be generalized in various ways by allowing it to take on arbitrary real values and to use various \textit{transfer functions} (also called an \textit{activation function}) other than the heavy-side step function. A transfer function is a function $g:\mathbb{R}\rightarrow \mathbb{R}$ that maps the weighted input of the neuron to its output. Popular transfer functions for neurons that take real numbers as activation values include $tanh$, the logistic function $\sigma(x) = \frac{1}{1+e^{-x}}$, and the rectified linear function: \[Relu(x)=   \left\{
\begin{array}{ll}
      x & \textit{if } x \geq 0 \\

      0 & \textit{if } x < 0 \\
\end{array} 
\right. \] The values of the neurons are then updated with the equation $x_i'=g_i(\sum\limits_j w_{j,i}x_j +b_i)$, where $g_i$ is the transfer function of the $i^{th}$ neuron. A \textit{state} of a neural network with $n$ neurons is a vector $(x_1,x_2,...,x_n)\in \mathbb{R}^n$ representing an assignment of values to each neuron. The \textit{state space}, denoted by $X\subseteq\mathbb{R}^n$, is the set of all such vectors that are allowed. We often restrict the state space to a proper subset of $\mathbb{R}^n$ even if the update equations are defined on the entirety of $\mathbb{R}^n$. A neural network, therefore, defines a \emph{dynamical system} on its state space, as follows:
\begin{equation}
\label{nnd}
    x^{t+1}=g(Wx^t+b),
\end{equation}
where $x^t$ is the vector representing the state of each neuron at time $t\in \mathbb{N}$, $g$ is the stack of transfer functions for each neuron, i.e. $g(x_1,...,x_n)=(g_1(x_1),...,g_n(x_n))$, $W$ is the weight matrix and $b$ is the vector of biases. This definition does not include some variants of neural networks. In particular, probabilistic neural networks, neural networks with delayed connections, and neural networks that operate in continuous time. We will cover probabilistic networks when we discuss probabilistic semantic encodings and for simplicity we will not discuss continuous time networks or networks with delayed connections, although our definition of semantic encoding can be applied to those cases too.  

In neuro-symbolic computing, the state of a network is meant to represent meaningful symbolic information. However, a neural network often has hidden neurons which increase the computational power but do not themselves represent relevant symbolic information. In these networks, a semantic encoding provides a semantic interpretation to the \textit{visible} neurons (typically input and output or bottleneck neurons). Furthermore, in some cases a network may require intermediate computational steps in order to implement a specific update function. To apply our definition of semantic encoding correctly in these cases, we will equip neural networks with an equivalence class on the state-space induced by its hidden neurons. This equivalence class defines the set of states of the network that are meaningfully distinct. We will also give a network an additional parameter called the \textit{computation time}, $t_c\in \mathbb{N}$, which indicates the number of times that Equation \ref{nnd} should be applied for a single update of the network's semantics. 
This leads to the following definition of a \textit{candidate network}:
\begin{definition}
\textit{Let $N$ be a neural network with $n$ neurons and let $X\subset \mathbb{R}^n$ be its state space. Given a partition of $\{1,2,...,n\}$ into two sets named the visible units $V\neq \emptyset$, and the hidden units $H$, a candidate network is a triple $(N,\sim_N,t_c)$, where $\sim_N$ is the equivalence relation on $X$ defined by $x\sim_N x'$ if and only if $x_i=x_i'$ for all $i\in V$, and $t_c\in \mathbb{N}$ is a positive integer. We write $N(x)$ to denote the result of updating the state, $x$, $t_c$ times according to Equation \ref{nnd}. and $N^k(x)$ is the result of updating the state $t_c\cdot k$ times. }
\end{definition}
\noindent If a neural network has no hidden units then all states of the network are semantically-relevant and each candidate network is $(N,=,t_c)$ for some $t_c$. 
 
In the following example and throughout the paper we will represent a neural network visually as in Figure \ref{nnfig:2} by representing neurons with a labeled graph node, connections $w_{ij}$ as a weighted edge from node $i$ to node $j$, and biases $b_i$ as a weighted edge with node $i$ as a target and no source. If a node does not have an edge for the bias then the neuron represented by the node has a bias of $0$. 
\begin{example}
\textit{
Consider the network in Figure \ref{nnfig:2}. Assume that all neurons take binary values and have the Heaviside step function as their activation function. As shown in the figure, all nodes have biases of $-0.5$. The state space is $\{(0,0,0),(0,0,1),(0,1,0),(0,1,1),(1,0,0),(1,0,1),(1,1,0),(1,1,1)\}$. If neurons $x_2$ and $x_3$ are considered to be hidden then our equivalence relation is $(x_1,x_2,x_3)\sim_N (x_1',x_2',x_3')$ iff $x_1=x_1'$. The equivalence classes are $\{(0),(1)\}$ where $(0)$ is the equivalence class containing all states in which $x_1=0$, and $(1)$ is the equivalence class containing all states in which $x_1=1$. 
Consider three candidate networks: $(N,\sim_N,1)$, $(N,\sim_N,2)$ and $(N,\sim_N,3)$. By simple calculation (i.e. propagation of activation through the network), in the first case $N((x_1,x_2,x_3))=(x_3,x_1,x_2)$, in the second case $N((x_1,x_2,x_3))=(x_2,x_3,x_1)$, and in the third case $N((x_1,x_2,x_3))=(x_1,x_2,x_3)$. This shows that updating the third candidate network is equivalent to the identity function. }
\end{example}
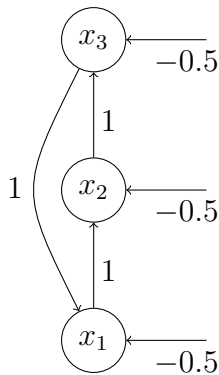
\begin{figure}

\centering
\begin{tikzpicture}

\useasboundingbox (0,0) rectangle (4,6);

 \node[shape=circle,draw=black, anchor=center,minimum size=0.8cm] (x_1) at (2,1){$x_1$};
 
  \node[shape=circle,draw=black, anchor=center,minimum size=0.8cm] (x_2) at (2,3){$x_2$};

  \node[shape=circle,draw=black, anchor=center, minimum size=0.8cm] (x_3) at (2,5){$x_3$} ;
  
  \coordinate (b_1) at (3.5,1);
  \coordinate (b_2) at (3.5,3);
  \coordinate (b_3) at (3.5,5);
 
\draw[->] (x_1) -- (x_2) node[ below, xshift=0.2cm, yshift=-0.8cm]{$1$};

\draw[->] (x_2) -- (x_3) node[ below, xshift=0.2cm,yshift=-0.8cm]{$1$};

\draw[->] (x_3) .. controls (1,3) .. (x_1) node[near start, above, xshift=-0.4cm,yshift=-0.9cm]{$1$};

 \draw[->] (b_1) -- (x_1) node[near start, below, xshift=0.0cm]{$-0.5$}; 
 \draw[->] (b_2) --  (x_2) node[near start, below,xshift=-0.0cm]{$-0.5$};
  \draw[->] (b_3) --  (x_3) node[near start, below,xshift=-0.0cm]{$-0.5$};

\end{tikzpicture}
\caption{A simple recurrent neural network with $3$ neurons}
\label{nnfig:2}
\end{figure}
\noindent
Candidate networks will be one of the two components being related by semantic encoding; the other being the knowledge-base of a logical system. From this point on, we will never have to work with a neural network that is not a candidate network. For brevity, we will often refer to a candidate network $(N,\sim_N,t_c)$ simply as a neural network and we will specify the hidden units and computation time. When the hidden units and computation time are not specified, the corresponding candidate network should be taken to be $(N,=,1)$.  

While the majority of neural networks that we discuss can be described as outlined above, not all the neural networks used in practice conform to the definitions. 
Feed-forward neural networks generally have their input state fixed rather than updated.  Many recently developed models, such as graph neural networks are not updated via a simple summation of weights, etc. Despite this, our framework can be applied to these other networks too. All that is necessary for our definitions to apply is the existence of a state space, $V=V_{visible}\times V_{hidden}$ and a mapping $N: V \rightarrow V$. All our illustrative examples will use neural networks as defined above. Examples from the literature that do not use neural networks as defined above will be explained in more detail.
 
Many neuro-symbolic techniques make use of \textit{stable state semantics}. In stable state semantics, the models of a logical system  are represented by the fixed points of a function. In order to make use of stable state semantics with neural networks, we must guarantee that the neural network will always converge to a stable state. This motivates the following definition.
\begin{definition}
\textit{Given a neural network, $N$, with state space $X$, a point $x \in X$ is stable if there exists $t>0$ such that for all $t'\geq t$, $N^{t'}(x)=N^t(x)$. }
\end{definition}
\noindent For many of the networks we investigate, given any initial state, the state of the network will eventually converge to a fixed point. This is a useful property for creating semantic encodings. It  means that we can build a semantic encoding by designing a network to have a specific set of fixed points. Note that all feed-forward networks have this property. In particular, a feed-forward network with $k$ layers will settle onto a fixed point after $k$ updates. To see this, first consider the values of the neurons in the input layer after the first time step. Without a change in the input, the state of the first layer is determined solely by the biases of the input neurons and is thus fixed after the first time step. Accordingly, the values of the neurons in the second layer do not change after the second time step, and so on until the values of the neurons in all layers are fixed. While this property of feed-forward networks is sometimes used in semantic encodings of logic programs, more often feed-forward networks are viewed as functions from the initial values of the input layer to the output layer. In these cases, we clamp the state of the first layer to the values of the input. We will make sure to distinguish between the two cases when they appear.  
 
As we will see in the following section, if a network settles to a fixed-point, then the states which carry the relevant semantic information are the stable ones. Otherwise, we will look to the states that satisfy weaker conditions, such as \emph{infinite recurrence}, as defined below.
\begin{definition}\textit{
Given a neural network $N$ with state space $X$, we say that $x\in X$ is infinitely recurring if there exists $x_0\in X$ such that for all $t>0$, $\exists t'>t$ such that $N^{t'}(x_0)=x$. }
\end{definition}
Notice that all stable states are infinitely recurring. If the state space of a neural network is finite, then after enough time, the state of the neural network will always be an infinitely recurring state. However, we also want our framework to accommodate networks with continuous state spaces and so there is one more case to consider, as follows.
 
\begin{definition}
\textit{
Given a neural network, $N$, with state space $X$, we say that $x\in X$ is a limit point of $N$ if there exists a sequence $\{N^{t_i}(x_0)\}_{i=1}^{\infty}$, $t_i<t_{i+1}$, with $\lim\limits_{i\rightarrow \infty} N^{t_i}(x_0)=x$. Call the set of limit points $X_{\textit{inf}}$.}
\end{definition}
It is easy to see that if $x$ is infinitely recurring then $x\in X_{\textit{inf}}$. Furthermore, if $X$ is finite then $X_{\textit{inf}}$ is exactly the set of infinitely recurring points. It will be the semantic interpretation of $X_{\textit{inf}}$ that determines whether or not a neural network encodes a knowledge-base. Our final requirement for neural networks is that they are stable in the sense that they always converge to $X_{inf}$.
\begin{definition}
\textit{A neural network, $N$, is stable if $\lim\limits_{t\rightarrow \infty} d(N^t(x),X_{inf})=0$ where $d(x,X_0)$ is the distance between a point and a set defined by $d(x,X_0)=\inf\limits_{x_0\in X_0} |x-x_0|$.}
\end{definition}
Generally speaking, most of the networks we examine will always settle on a fixed point. However, we also include the cases where the network converges to a cycle or a general limit point. Generalizing our definitions to networks that do not have this property (for example, networks which exhibit chaotic behaviour) will be left as future work. 

\subsection{Logical Systems}
Logical systems are expected to capture the fundamental properties of logic by abstracting away the meaning of sentences. This allows logical deduction to be seen as a relationship between sentences determined by their structural properties. A logical system is thus comprised of a \textit{language} and some notion of entailment which determines whether one sentence follows logically from a \textit{knowledge-base}, which we define as a set of sentences. In order to keep our definitions sufficiently general, we define a logical system as a pair $\mathcal{S}=(\mathcal{L},\vdash_{\mathcal{S}})$ where $\mathcal{L}$ is a computable language and $\vdash_{\mathcal{S}}$ is a computable relation on $2^\mathcal{L}$. Usually, $\vdash_{\mathcal{S}}$ is defined either as a deductive system, in which a set of predefined rules and axioms are used to compute, given $L,L' \subseteq \mathcal{L}$, whether or not $L\vdash_{\mathcal{S}} L'$, or $\vdash_{\mathcal{S}}$ is defined semantically by a set of interpretations which determine a truth-value for every $L\subseteq \mathcal{L}$. Because we are looking at semantic encodings of logical systems into neural networks, we will
use the following as our formal definition of a logical system.
\begin{definition}
Given a language $\mathcal{L}$, and a set $\mathcal{M}\subseteq \{f | f:2^\mathcal{L}\rightarrow \{0,1\}\}$, a logical system is a pair $\mathcal{S}=(\mathcal{L},\vdash_{\mathcal{S}})$, where $\vdash_{\mathcal{S}}$ is the relation defined by $L\vdash_{\mathcal{S}}L'$ iff for all $M\in \mathcal{M}$, if $M(L)=1$ then $M(L')=1$.
\end{definition}
The functions $M\in \mathcal{M}$ are intended to interpret the abstract sentences in $\mathcal{L}$ in such a way that the truth of each knowledge-base $L$ can be determined. If $M(L)=1$ then $L$ is \emph{true} in $M$ and if $M(L)=0$ then $L$ is \emph{false} in $M$. $M$ is thus referred to as an \textit{interpretation} of the logical system. If $M(L)=1$ we say that $M$ is a \textit{model} of $L$. We use $\mathcal{M}_L$ to denote the set of models of $L$. When we refer to a logical system, we will assume that $\vdash_{\mathcal{S}}$ is defined by some set of interpretations $\mathcal{M}$. Because we are only looking at entailment from a semantic point of view, we will use the symbol $\vDash_{\mathcal{S}}$ in place of $\vdash_{\mathcal{S}}$ which is more commonly reserved for entailment defined by deductive rules. Because $\vDash_{\mathcal{S}}$ is completely defined by $\mathcal{M}$, we will refer to a logical system as $\mathcal{S}=(\mathcal{L},\mathcal{M})$. 
 
In many logical systems used in AI, sentences are often augmented with additional information in the form of a \textit{label}. A label is meant to confer additional information about a sentence that is not conveyed in the language itself. Examples of labels include confidence levels, target truth-values or timestamps. A formal framework to define labelled deductive systems can be found in \cite{LDS}. For simplicity, we stick with the standard previously established definition of a logical system, but our definition of semantic encoding could also be applied to the logic framework of Labelled Deductive Systems (LDS). We conclude this section with an example using a propositional fuzzy logic, later to be encoded into a neural network using Logic Tensor Networks (LTN) \cite{LTN}.
 
\begin{example}
\textit{
Let $\mathcal{L}_p$ be the language of propositional logic constructed from the set of variables $\mathcal{X}=\{X_1,X_2,...\}$ and the logical symbols $\{\neg,\vee\}$. Define $\mathcal{L}_f$ as the set of sentences of the form $[a,b] : l$, where $a,b\in [0,1]$, $a\leq b$ and $l\in \mathcal{L}_p$. Given a function $M_t:\mathcal{X}\rightarrow [0,1]$, we define a function $\hat{M}_t:\mathcal{L}_p\rightarrow [0,1]$ recursively:
\begin{equation*}
    \begin{aligned}
        \hat{M}_t(X_i)=M_t(X_i) \\
        \hat{M}_t(\neg l)=1-\hat{M}_t(l) \\
        \hat{M}_t(l\vee l')=\max(\hat{M}_t(l),\hat{M}_t(l'))
    \end{aligned}
\end{equation*}
Given a sentence $[a,b]:l \in \mathcal{L}_f$, if $\hat{M}_t(l)$ is in the interval $[a,b]$ then we say that $\hat{M}_t$ satisfies the sentence. We construct an interpretation, $M$, from $\hat{M}_t$ by defining $M(L)=1$ iff $\hat{M}_t$ satisfies all sentences in $L$. The set of interpretations is in a one-to-one correspondence with the set of mappings $\hat{M}_t$. To demonstrate entailment, consider the knowledge-base $L=\{[0,0.1]: A,\  [0.4,0.5]: A \vee B\}$. Because all models of $L$ must satisfy $\hat{M}_t(A)\in [0,0,1]$, and $\hat{M}_t(A\vee B) \in [0,4,0.5]$, they must also satisfy $\hat{M}_t(B)\in [0.4,0.5]$ and so $L\vDash_\mathcal{S} \{[0.4,0.5]:B\}$.
}
\end{example}

With the above definitions of neural networks and logical systems we can formally define what we mean by a semantic encoding between the two. To give a preview, the idea will be to represent a neural network with a knowledge-base in a logical system. This is done by mapping each state of the neural network to interpretations of the logical system. By aggregating the interpretations represented by the limit points of the network, we associate the network with a set of interpretations for the logical system. If every element of this set is a model of a knowledge-base then we call the neural network a neural model of the knowledge-base. If this set of interpretations fully determines $\vDash_{\mathcal{S}}$ for the knowledge-base then we call the neural network a semantic encoding of the knowledge-base. We formalize this in the next section. First, we take a brief look at the various approaches to neuro-symbolic computation that will be unified within the same framework of semantic encoding.

\subsection{Neuro-symbolic Integration}

Approaches to neuro-symbolic integration generally fall into three categories: neural encoding, rule extraction and hybrid systems. While the latter two are interesting in their own right, it is neural encoding that is the topic of this paper. Neural encoding is the process of representing a symbolic system by a neural network either through its architecture and weights or through a loss function. The neural network is then free to be trained on additional examples with the hope that the symbolic background knowledge representation will guide the network to find solutions adhering to the conditions written into the network by the knowledge.

The overall challenge for neural encoding is how to map the many forms of knowledge representation that have been proven to be useful in symbolic AI into a neural network. These include temporal, nonmonotonic, epistemic, ontological knowledge as well as normative reasoning and argumentation systems \cite{NeSyBook}. It requires answering the question of how expressive neural networks are \cite{fibring,EpistemicProblems}. In practice, it implies striking a balance between encoding rich forms of knowledge representation such as first-order, many-valued and higher-order logics, yet maintaining the ability of neural networks to perform efficiently as a model for learning and computation. 

A prerequisite for neuro-symbolic computing, before the above challenges can be addressed, is to answer the question: what does it mean for a symbolic system to be encoded correctly in a neural network? Many competing answers have been given for this question, generally within the context of introducing a new encoding algorithm. The first methods to address this question were the \textit{core} methods \cite{CORE,CILP}, which showed that a neural network can implement the \textit{least fixed point operator} of a logic program.\footnote{The least fixed point operator, $T_P$, of a logic program, $P$, is a mapping between interpretations of the language of $P$ defined as $T_P(M)=\{A | \exists (A\leftarrow X_1\wedge X_2 \wedge...\wedge X_k) \in P, X_1,X_2,...,X_k \in M\} $.} 
As a result, a knowledge-base that can be described as a logic program can be encoded in a neural network directly by setting the weights so that the network can be shown to implement the semantics of that knowledge-base. Other methods at the time used energy based models and claimed that a neural network was equivalent to a logical system when the minima of the network's energy function corresponded to the models of a knowledge-base \cite{penalty,Trann:2016}. More contemporary methods tend to be approximate, some use fuzzy logic operators to regularize the loss function while others use probabilisitc measures \cite{LTN,deepproblog}. In these cases, the claim is that a neural network encodes a knowledge-base when this loss tends to zero. Other results have claimed logical equivalence between graph neural networks and fragments of first order logic, or that transformers are equivalent to first order logic extended with majority quantifiers \cite{transformer_equiv,GNN:semantic_equiv}. All of these approaches are obviously related in the sense that they connect neural networks with logical systems, but comparing them is difficult as the logic and networks vary, in some case being vastly different, and the manner in which they are related (fixed-points, energy valleys, marginal probabilities, etc) may appear to be incompatible with one another.


We will show that all the methods mentioned above and variations thereof have the same underlying components and implicit definition of equivalence. Our definition of semantic encoding will make this equivalence explicit. In practice, semantic encoding will offer a set of definitions of the components and conditions that must be satisfied in order to claim that a knowledge-base is encoded correctly by a neural network. There are numerous advantages to having a unifying framework. The most immediate benefit is that it allows us to compare existing methods in a systematic way. It is not immediately obvious how, e.g. LTN relates to the CORE method, but our framework will show that the primary difference is in the choice of aggregation and encoding functions, as defined in the next section. As a result, future encoding methods will be more easily situated within the context of the existing work by identifying explicitly their choice of components (mapping, aggregation and encoding, as illustrated in Table \ref{table:semantic_encoding_comparison}). While almost all encodings used today rely on \textit{stable-state semantics}, that is, they assume that the neural network converges to a stable state, our framework generalizes this notion by allowing for recurrent networks with more complex dynamics to qualify as a semantic encoding. This will be shown to offer a straightforward method for extending existing techniques to new classes of encoding that will retain many of their original properties.

The final advantage of a unifying framework is that it opens up the possibility of developing a general theory of semantic encoding. The potential long-term benefit of this is to offer the tools that can help organize the research around the properties, constraints and results of entire sets of semantic encodings, rather than the properties, constraints and results that apply only to a single method or small family of methods. We will discuss this in more detail at the end of the next section. 

\section{The Semantic Encoding Framework}

We now formalize the concept of semantic encoding. 
In a semantic encoding, each state of a neural network represents information about a set of interpretations of a logical system. For example, in the common \textit{neurons-as-atoms} paradigm, each neuron represents a logical atom and the value of the neuron determines the truth-value of the atom. Each state of the neural network thus represents an assignment of truth-values to a set of atoms, and thus represents an interpretation or a set of interpretations of a logical system. 
In a distributed setting, the situation is similar but with each atom represented by a set of neurons and the corresponding interpretations represented by the neurons' activation values.
In most semantic encodings in the literature, the state of the neural network is sufficient to determine whether or not the corresponding interpretations are models of the knowledge-base. This is the case when the state of the neural network determines the truth-value of every atom contained in the knowledge-base. In some more recent semantic encodings, however, the state of the neural network only represents some of the information required to determine whether or not the interpretations are models of a target knowledge-base. For example, in Logic Tensor Networks \cite{LTN}, the state only determines the truth-value of \textit{some} of the atoms in the target knowledge-base. This can be viewed as a state mapping to the set of interpretations in which the atoms represented by the state have their truth values agree with their assignments in the state.
 
For either case, the central component is a mapping $i:X\rightarrow 2^\mathcal{M}$, where $X$ is the state space of a neural network and $2^{\mathcal{M}}$ is the set of interpretations of a logical system. Because we assume that the semantic information is encoded in the state of the visible units of the network, if $x_1$ and $x_2$ have an identical visible configuration then their corresponding sets of interpretations should also be identical. Furthermore, we would like $i$ to represent a `natural' mapping. That is, $i$ should not be an arbitrary association of neural states and interpretations, but should instead satisfy certain commonsense constraints that arise from the particular networks and logical systems being considered. An example of this might be a notion of continuity of $i$. That is, small changes in the state of the network should represent correspondingly small changes in the interpretations. Because the exact constraints on $i$ may depend on the context, we impose no hard requirements in our definition. We simply assume that $i$ belongs to a set of functions that we call \textit{candidate maps} that will satisfy certain constraints. A good example of a set of candidate maps are the previously discussed neurons-as-atoms mapping, which we denote by $I_{NAT}$. As mentioned, these are mappings which associate states of the neural network to interpretations by identifying atoms with single neurons. More on the set of candidate maps will be discussed in the next section. For now, we merely assume that there is some set of candidate maps and use them to define a relationship between states of the neural network and interpretations of a logical system. We formalize this with the following definition.

\begin{definition}
\textit{Given a logical system $\mathcal{S}=(\mathcal{L},\mathcal{M})$ and a neural network $N$ with state space $X$, $I\subset \{i| i:X\rightarrow 2^{\mathcal{M}}\}$ is a set of candidate maps if for all $i\in I$, we have that $x_1 \sim_N x_2$ implies that $i(x_1)=i(x_2)$. We call $i$ an encoding function.}
\end{definition}

\noindent Our ultimate goal is to be able to encode a knowledge-base into a neural network in a way that makes the two equivalent. The mapping $i$ gives us a way of associating states of a neural network with interpretations of a logical system, but we are yet to define how the network encodes a knowledge-base. The most common approach is to design a neural network in which the interpretations represented by the stable states are models of the knowledge-base. The neural network is expected, therefore, to always converge to a model of the knowledge-base. Intuitively, we can think of this as revising the beliefs about the true state of the world until they satisfy a set of constraints which are assumed to be true. If every model of the knowledge-base is represented by the network in this way, or at least enough models to determine the semantics of the knowledge-base, then we call the network a semantic encoding of that knowledge-base. 
 
How do we generalize this to networks which may exhibit periodic or other, more complicated, dynamical behaviour? To answer this, consider the case of the Necker Cube. The Necker Cube is an optical illusion in which there are two equally valid interpretations of an image; one in which the cube is extending outward and one in which the cube is extending inward. After looking at the Necker cube for long enough, it is common for our interpretations to switch back and forth between the two valid interpretations. With this in mind, we will define a neural encoding of a knowledge-base $L$ as a network that converges to states corresponding to the (possibly many) models of $L$. In the Necker cube example, this would mean the network converges to one of the two interpretations before switching between them in a cycle.
 
The final thing we need to define is a method of aggregating the models represented by different states of the network. Each state of the network represents a set of interpretations, and a set of states represents a set of sets of interpretations. In order to define a set of interpretations corresponding to a network, we must have some way of combining the sets of interpretations given by different states of the network. For this reason, in addition to $i$, we assume that we are given an aggregation function, $Agg$. In all examples to follow, this will either be \emph{set union} or \emph{intersection}, but we keep $Agg$ generic in the definition below.
\begin{figure}
    \centering
    \includegraphics[scale=0.6]{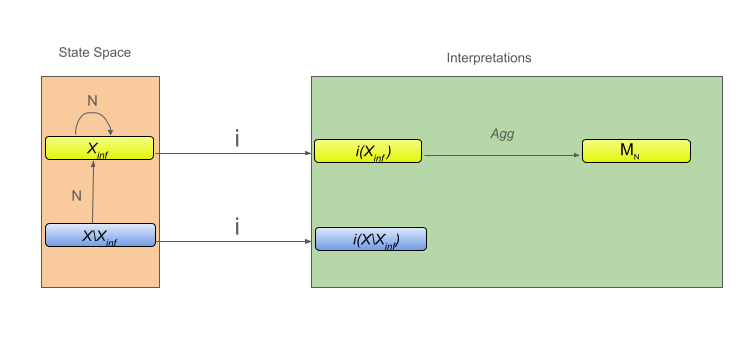}
    \caption{A block diagram of a semantic encoding. The stable states of a neural network are mapped to sets of interpretations which are aggregated into a single set of interpretations, $\mathcal{M}_N$. If these interpretations are models of a knowledge-base then the neural network is said to be a \textit{neural model} of the knowledge-base. If these interpretations represent all models of the knowledge base (or a sufficient number of them to determine the logical entailment relation) then the neural network is said to be a \textit{semantic encoding}.}
    \label{fig:semantic_encoding}
\end{figure}
\begin{definition}
\label{semanticneuralmodel}
\textit{ Let $\mathcal{S}=(\mathcal{L},\mathcal{M})$ be a logical system, $N$ a neural network with state space $X$, and let $L$ be a knowledge-base of $\mathcal{S}$. Given an encoding function $i\in I$, $i:X\rightarrow 2^{\mathcal{M}}$, and an aggregation function, $Agg:2^{2^{\mathcal{M}}}\rightarrow 2^{\mathcal{M}}$, we define:
\begin{itemize}
    \item Let $\mathcal{M}_N=Agg(\{i(x) | x\in X_{\textit{inf}}\})$. $N$ is called a \emph{neural model} of $L$ under $I$ and $Agg$ if $\emptyset \subset \mathcal{M}_N \subseteq \mathcal{M}_L$.
    \item $N$ is called a \emph{semantic encoding} of $L$ under $I$ and $Agg$ if it is a neural model of $L$ under $I$ and $Agg$ and $L\vDash_{\mathcal{S}} L'$ iff $\mathcal{M}_N\subseteq \mathcal{M}_{L'}$.
    \item A set of neural networks, $\mathcal{N}$, and the logical system, $\mathcal{S}$, are \emph{semantically equivalent} under $I$ and $Agg$ if every knowledge-base of $\mathcal{S}$ has a semantic encoding under $I$ and $Agg$ to a neural network in $\mathcal{N}$ and all neural networks in $\mathcal{N}$ are semantic encodings under $I$ and $Agg$ of some knowledge-base $L$.
\end{itemize}
}
\end{definition}

\noindent We say that a neural network is a neural model of $L$ if, for any initial state, given enough time, the state of the neural network encodes information about models of $L$. How we combine this information to determine which specific models of $L$ the network represents depends on the aggregation function. In all the examples to follow, $Agg$ will be a union if $i$ maps each state in $X_{\textit{inf}}$ to a set of models of $L$, and $Agg$ will be an intersection otherwise. However, as mentioned, we leave open the possibility of alternative aggregations. In a semantic encoding, the models of $L$ represented by $N$ are sufficient to fully determine the semantics of $L$. $L$ may have models not contained in $\mathcal{M}_N$, but the validity of $L\vDash_{\mathcal{S}} L'$ only depends on the models represented by the network $N$. This is visualized in Figure \ref{fig:semantic_encoding}.
It may be useful to view the states of the network as representing beliefs about the world with $i$ mapping to interpretations which satisfy those beliefs. The beliefs of a neural network are encoded in states that it converges to over time. A neural model of $L$ is therefore a neural network whose beliefs satisfy $L$. Note that if $\mathcal{M}_N=\emptyset$ then the network trivially satisfies the condition for any knowledge-base. This is the case in which the neural network has no beliefs. This can happen if the beliefs held by the relevant stable states are contradictory and $Agg=\cap$. We can thus interpret a neural model of $L$ as a neural network that represents some set of beliefs satisfying $L$. 
 
A semantic encoding is a neural model that can be used to perform logical inference. In a semantic encoding, the beliefs of a network, represented by $\mathcal{M}_N$, not only satisfy $L$, but if they satisfy $L'$ then $L\vDash_{\mathcal{S}} L'$. This can be most easily shown by proving that $\mathcal{M}_N=\mathcal{M}_L$, which is how we generally prove that a neural network is a semantic encoding throughout this paper. However, in certain cases, this condition would be impossible to satisfy for any neural network, encoding function and aggregation function. In particular, in first-order logic, the Lowenheim-Skolem theorem states that a knowledge-base that admits a model of infinite cardinality has models of arbitrary cardinality \cite{intrologic}. This means that the collection of all models for many knowledge-bases in first-order logic is not a set, but a proper class, whereas by definition $M_N$ must be a set.  
 
Next, we illustrate the above definitions with several examples.
\begin{figure}

\centering
\subfloat[Network architecture]{
\centering
\begin{tikzpicture}

\useasboundingbox (0,0) rectangle (4,4);

 \node[shape=circle,draw=black, anchor=center,minimum size=0.8cm] (y_1) at (0,2){$x_A$};

 \path (y_1) edge [in=50,out=130,loop] node[above]{$-1$} (y_1) ;
 
  \node[shape=circle,draw=black, anchor=center, minimum size=0.8cm] (y_2) at (4,2){$x_B$};
  \path (y_2) edge [in=50,out=130,loop] node[above]{$-1$} (y_2);
 
 \draw[->] (y_2) .. controls (2,2.25) .. (y_1) node[near end, below,yshift = 0.5cm,xshift=0.5cm]{$-1.5$};
  \draw[->] (y_1) .. controls (2,1.75) .. (y_2)  node[near end, below,yshift = -0.1cm,xshift=-0.8cm]{$-1.5$};
  
  \coordinate (b_1) at (-1,2);
    \coordinate (b_2) at (5,2) ;
  
 \draw[->] (b_1) -- (y_1) node[near start, below, xshift=0.1cm]{$2$}; 
 \draw[->] (b_2) --  (y_2) node[near start, below,xshift=-0.1cm]{$2$};
\end{tikzpicture}
}         
\hfill
\subfloat[State transition diagram and corresponding models]{\includegraphics[scale=0.5]{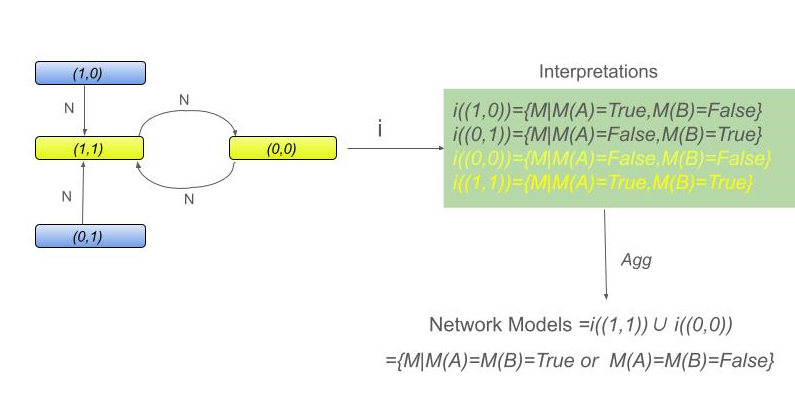}}
\caption{A recurrent neural network that semantically encodes the propositional knowledge-base $\{(A \wedge B) \vee (\neg A \wedge \neg B)\}.$ \label{nnfig:3}}
\end{figure}
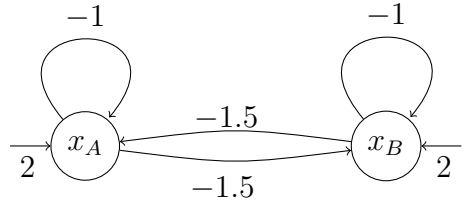
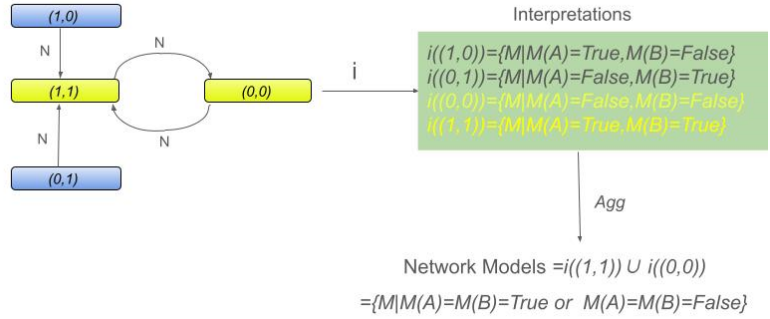
\begin{example}
\label{exmp:1}
\textit{Consider the propositional knowledge-base $L =\{(A \wedge B) \vee (\neg A \wedge \neg B)\}$. The models are all propositional variable assignments with either $A=B=True$ or $A=B=False$. We will give a semantic encoding for this knowledge-base under $I_{NAT}$ with $Agg=\cup$. Take a network with two neurons, each with a bias of $2$, self connections of $-1$, and a symmetric connection between them of $-1.5$ (see Figure \ref{nnfig:3}). This network has no hidden neurons and an update time of $1$, i.e. the candidate network is $(N,=,1)$, and the transfer function for each neuron is the heavyside step function. There are 4 states of this network, $(0,0),(0,1),(1,0),(1,1)$. We map these states to propositional truth assignments by identifying the first neuron with the variable $A$ and the second with $B$. Given a state, if a neuron has activation value $1$ (respectively $0$), the corresponding variable gets assigned $True$ (respectively $False$) as usual. For example, $(0,1)$ maps to the set of propositional truth assignments with $A=False$ and $B=True$. Note that this mapping is an example of a neurons-as-atoms mapping, which we will define in the next section, and thus we have $i\in I_{NAT}$. 
Under this mapping, the states corresponding to the models of the knowledge-base are $(0,0)$ and $(1,1)$. Calculating each state transition of the neural network reveals that for any initial state, the network will converge to the cycle $(0,0)\rightarrow (1,1)\rightarrow (0,0)$ meaning that $X_{inf} = \{(0,0),(1,1)\}$. We have $i((0,0))=\{M | M(A)=False,M(B)=False, M \in \mathcal{M}\}$ and $i((1,1))=\{M | M(A)=True, M(B)=True, M \in \mathcal{M}\}$, so $\mathcal{M}_N=Agg(i((0,0)),i((1,1)))=i((1,1))\cup i((0,0)) = \mathcal{M}_L$. Because $\mathcal{M}_N\subseteq \mathcal{M}_L$, $N$ is a neural model of $L$ under $I_{NAT}$. Furthermore, because $\mathcal{M}_N=\mathcal{M}_L$, $L\vDash_{\mathcal{S}} L'$ iff  $\mathcal{M}_N\subseteq \mathcal{M}_L$ and $N$ is a semantic encoding of $L$ under $I_{NAT}$ and $\cup$. This is illustrated in Figure \ref{nnfig:3}(b) in which the state transition diagram is shown on the left with states in $X_{inf}$ shown in yellow. Each state is mapped to a set of interpretations on the right with the images of $X_{inf}$ again highlighted in yellow. These interpretations are passed to the aggregation function, in this case union, to arrive at the final set of interpretations, which are the models of the knowledge-base. } 
\end{example}

In the above example, the models of $L$ were fully determined by the truth assignments of $A$ and $B$. Because each state of the neural network fully determined $A$ and $B$, each individual state of $X_{inf}$ represented a set of models of $L$ and the complete set of models represented by $N$ was the union of the sets of models represented by each state in $X_{inf}$. Now we look at an example in which each state of the network represents a truth assignment to a subset of the atoms in the knowledge-base. In this case and others like it, set intersection is used to combine the information represented in $X_{inf}$. Furthermore, we introduce the set of candidate maps $I_{DAT}$, which are maps, $i$, that represent atoms using distributed patterns of activity. $I_{DAT}$ will be formally defined alongside $I_{NAT}$ in the next section.

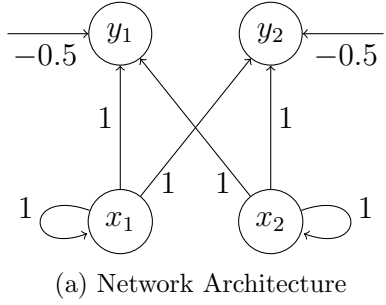
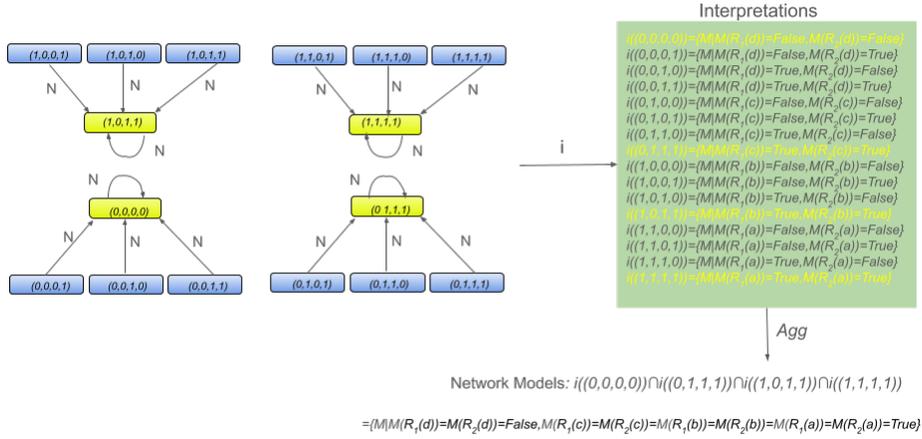
\begin{figure}
\centering

\subfloat[Network Architecture]{\begin{tikzpicture}[scale=1]

\useasboundingbox (0,0) rectangle (4,4);

 \node[shape=circle,draw=black, anchor=center,minimum size=0.8cm] (a) at (1,0.5){$x_1$};

 \path (a) edge[in=200,out=160,loop]  node[ above, xshift=-0.2cm,yshift=-0.1cm]{$1$} (a);
  \node[shape=circle,draw=black, anchor=center,minimum size=0.8cm] (b) at (3,0.5){$x_2$} ;

  \path (b) edge [in=-20,out=20,loop] node[ above,xshift=0.2cm,yshift=-0.1cm]{$1$} (b);

  \node[shape=circle,draw=black, anchor=center,minimum size=0.8cm] (c) at (1,3){$y_1$};

  \node[shape=circle,draw=black, anchor=center,minimum size=0.8cm] (d) at (3,3){$y_2$};
  
  \coordinate (b_1) at (-0.5,3);
  \coordinate (b_2) at (4.5,3);
 
\draw[->] (b_1) -- (c) node[ above, xshift=-1cm, yshift=-0.6cm]{$-0.5$};

\draw[->] (b_2) -- (d) node[ above, xshift=1cm, yshift=-0.6cm]{$-0.5$};


\draw[->] (a) -- (c) node[ above, xshift=-0.2cm,yshift=-1.4cm]{$1$};


 \draw[->] (a) -- (d) node[near start, below, xshift=0.0cm]{$1$}; 
 \draw[->] (b) --  (c) node[near start, below,xshift=-0.0cm]{$1$};
  \draw[->] (b) --  (d) node[ above,xshift=0.2cm,yshift=-1.4cm]{$1$};
  

\end{tikzpicture}
}
\hfill
\subfloat[State transition diagram and corresponding models]{\includegraphics[scale=0.5]{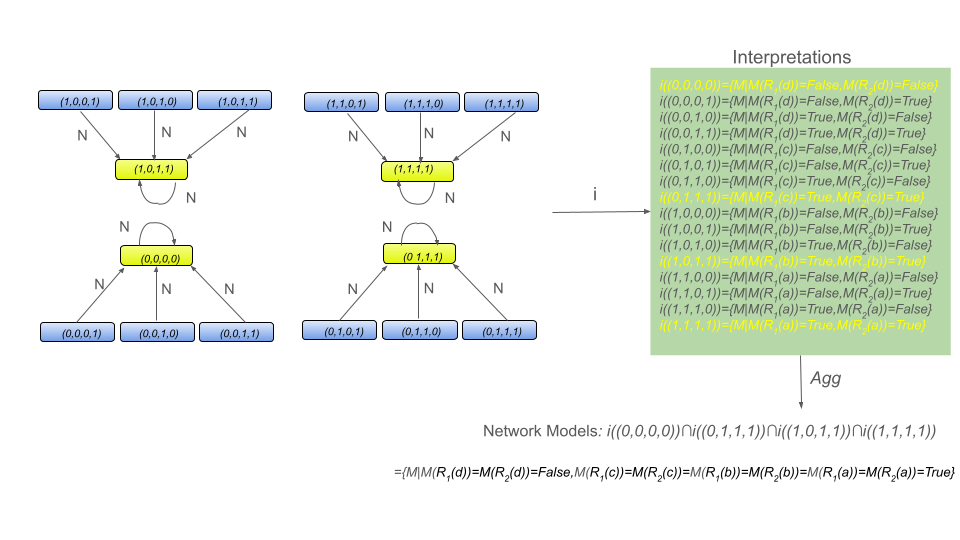}}
\caption{A neural network that semantically encodes the first-order knowledge-base $L=\{\forall x.(R_1(x)\leftrightarrow R_2(x)), R_1(a),R_1(b), R_1(c),\neg R_1(d)\}$ under $I_{DAT}$ and $Agg=\cap$.}
\label{nnfig:4}
\end{figure}

\begin{example}
\label{excsem}
\textit{Take a first order language, $\mathcal{L}$, consisting of a countably infinite number of variables, constant symbols $a,b,c,d$, binary relations $R_1,R_2$, and no function symbols. Define a logical system $\mathcal{S}=(\mathcal{L},\mathcal{M})$ where the interpretations are first-order structures over the Herbrand Universe of $\mathcal{L}$ (assignments of truth values to all atoms of the form $R(\xi)$ where $R\in \{R_1,R_2\}$ and $\xi \in \{a,b,c,d\}$). Take the knowledge-base:
\begin{displaymath}
L=\{\forall x.(R_1(x)\leftrightarrow R_2(x)), R_1(a),R_1(b), R_1(c),\neg R_1(d)\}.
\end{displaymath}
We give a semantic encoding of $L$ under $I_{DAT}$ with $Agg=\cap$. 
Take a neural network, $N$, with four binary neurons, $x_1,x_2,y_1,y_2$. Define the encoding function, $i$, by mapping $i(x_1,x_2,y_1,y_2)$ to the set of interpretations of $\mathcal{S}$ satisfying $R_1(\xi(x_1,x_2))=g(y_1)$ and $R_2(\xi(x_1,x_2))=g(y_2)$ where $g(0)=False$, $g(1)=True$  and $\xi(x_1,x_2)$ is defined by: 
\[\xi(x_1,x_2)=   \left\{
\begin{array}{ll}
      a: & x_1=x_2=1 \\
      b: & x_1=1,x_2= 0 \\
      c: & x_1=0,x_2=1 \\
      d: & x_1=0,x_2=0 \\
\end{array} 
\right. \]
Referring back to the desired knowledge-base, we can see that the network should implement $y_1=y_2= x_1 \vee x_2$. This is achieved by the following network: $x_1$ and $x_2$ have $0$ bias, self-connections of $1$, and a linear activation function; the weights are all $1$ ( $w_{x_1,y_1}=w_{x_1,y_2}=w_{x_2,y_1}=w_{x_2,y_2}=1$), and $y_1$ and $y_2$ have bias $-0.5$ and a step activation function (see Figure $\ref{nnfig:4}$). It is not difficult to calculate that $X_{inf}=\{(0,0,0,0),(0,1,1,1),(1,0,1,1),(1,1,1,1)\}$ and that $\cap_{x\in X_{inf}} i(x)$ consists of a single interpretation $M=\{R_1(a)=True,R_1(b)=True,R_1(c)=True,R_1(d)=False,R_2(a)=True,R_2(b)=True,R_2(c)=True,R_2(d)=False\}$ which is the unique model of $L$. Because $N^t(x)\in X_{inf}$ for all $t>1$, $N$ is a neural model of $L$ under $I_{DAT}$ and $Agg=\cap$. Because $\mathcal{M}_N=\mathcal{M}_L$, the network is a semantic encoding of $L$ under $I_{DAT}$ and $Agg=\cap$. We illustrate this in Figure $\ref{nnfig:4}$(b) in the same manner as the previous example. Notice the differences: the aggregation function is $Agg=\cap$, meaning that the final set of interpretations of the network is the intersection of the four sets of models mapped to by the stable states of the network, resulting in a unique model.
}
\end{example}

The ability to encode a knowledge-base into a neural network depends mostly on the particular set of candidate maps. With no restrictions on the candidate maps the existence of a semantic encoding is trivial, but with strong restrictions, such as $I_{NAT}$, there are important logical systems which cannot be represented by neural networks. Next, we address the question of this mapping in more detail.
\subsection{The Set of Candidate Mappings: Practical Considerations}
As we have seen, the most important component of a semantic encoding is the mapping from the state space of a neural network to interpretations of logical systems. This mapping determines how semantic information is encoded in the neural network and as such we expect it to make some amount of intuitive sense. In addition, developing neuro-symbolic encodings should not only be a theoretical exercise, but also provide methods for advancing the capabilities of AI systems. An arbitrary mapping may be impractical or not make intuitive sense. What constitutes a practical and intuitive candidate mapping depends on the network, the logical system, and the intended purpose of the semantic encoding. Next, we discuss several sets of candidate mappings and their limitations. The most common set is $I_{NAT}$, in which the truth-value of each atom only depends on a single neuron. This can be generalized to $I_{DAT}$, in which each atom is given a distributed representation. These are formally defined next, but before we do that, let us consider semantic encodings with unrestricted mappings. 
In the most general case, which we will refer to simply as $I_{total}$, the only requirement on $i$ is that $x\sim_N x'$ implies $i(x)=i(x')$. This means that there are no strong restrictions on the association of neural states with interpretations. Under this condition, we can map states to interpretations in an arbitrary way and thus the existence of a semantic encoding of a knowledge-base $L$ under $I_{total}$ depends only on the cardinality of the set of models of $L$ and the set of states $X_{\textit{inf}}$. For example, any knowledge-base can be semantically encoded into a neural network with a single stable state under $I_{total}$ by choosing $i$ to map the stable state to the set of models of the knowledge-base and mapping every other state to an arbitrary set of interpretations of the language.  

\noindent Semantic encodings under $I_{total}$ are of little practical use as they do not say anything about \textit{how} interpretations can be encoded into the state space of a neural network. The potential for arbitrary associations between interpretations and neural states leaves open the possibility of neural networks having too large a representation capacity. This is a problem that has been brought up in cognitive science. Putnam famously showed that, given any physical system, one can find a way to identify the states of the physical system with the states of an arbitrary automaton in such a way as to make the two equivalent in the sense that the physical system implements the automaton. This means that, mathematically, every physical system is an implementation of every automaton, making the identification meaningless \cite{putnam_cogsci}. To avoid this problem, we assume that the identification of neural states with interpretations preserves some kind of structure between the logical system and neural network. Exactly what this entails depends on the structure of the logical system being encoded. Because of this, we have not given any strict requirements for $i$ in our definition of semantic encoding. 

Let us now go over the restrictions that have been most commonly used for semantic encodings, $I_{NAT}$ and its generalization $I_{DAT}.$ 
Although we gave an abstract definition of a logical system, in practice, interpretations determine the truth-value of a sentence based on an assignment of truth-values to a set of atoms. An atom is the smallest unit to which an interpretation can assign a truth-value. Whether or not a knowledge-base is satisfied by a model depends solely on how a model assigns truth-values to the atoms. $I_{NAT}$ is a set of candidate mappings in which the truth-value of an atom is determined by a single corresponding neuron. Because atoms are left undefined in our definition of a logical system, we give an abstract characterization of them in the following definitions by associating the interpretations of a logical system with maps from a set of atoms to truth values. 
\begin{definition}
\textit{Let $\mathcal{S}=(\mathcal{L},\mathcal{M})$ be a logical system. Let $\mathcal{M}_0$ be a subset of interpretations that are defined by assignments of truth-values to atoms (i.e. there exist sets $A$ and $\mathbb{T}$, referred to as a set of atoms and truth-values, respectively, and a bijection $f:\mathcal{M}_0\rightarrow \mathbb{T}^{A}$). Let $i:X\rightarrow 2^{\mathcal{M}_0}$ be an encoding function of a neural network $N$ with $n$ visible neurons and state space $X\subseteq \mathbb{R}^n$. We say that $i\in I_{NAT}$ if there exist mappings $r:A'\rightarrow \{1,2,...,n\}$ with $ A'\subseteq A$ and $g:\mathbb{R}\rightarrow \mathbb{T}$ that satisfy the following properties: $r$ is bijective and if $M\in i(x)$ and $r(Q)=j$ then $f(M)(Q)=g(x_j)$ where $x_j$ is the value of the $j^{th}$ visible neuron. 
}
\end{definition}
\noindent Under $I_{NAT}$, a neuron represents an atom and the value of the neuron determines the truth-value of the corresponding atom. We have already seen multiple examples of encodings under $I_{NAT}$. Most of the semantic encodings of logic programming have been done under $I_{NAT}$. Unfortunately, $I_{NAT}$ is far too restrictive for a network to encode all interesting knowledge-bases, in particular in first-order logic, as has been discussed in \cite{neurorsym}. Under $I_{NAT}$, a knowledge-base that references $n$ atoms will require a network to have at least $n$ neurons to encode that model. If there are an unbounded number of atoms in the models of $L$, such as is the case when an interpretation in a first-order language has an infinite domain, no neural network can semantically encode $L$ under $I_{NAT}$. 

It has been argued that symbolic information should not be encoded in single neurons but rather as distributed patterns of activity occurring across many neurons \cite{proper_connectionism}. Because of this, the realization of fully-distributed logical reasoning within a neural network has been a central aim of neuro-symbolic computing \cite{LTNs,guha14a}. Taking inspiration from such work, we provide a generalization of $I_{NAT}$ that allows for atoms to be represented by distributed patterns of activity across the neural network. We call this set of candidate maps $I_{DAT}$.
\begin{definition}
\label{def:IDAT}
\textit{
Let $\mathcal{S}=(\mathcal{L},\mathcal{M})$ be a logical system with a subset of interpretations, $\mathcal{M}_0$, that are defined by $f:\mathcal{M}_0\rightarrow \mathbb{T}^A$. Let $N$ denote a neural network with $n$ visible neurons with labels $\{1,...,n\}$ and state space $X\subseteq \mathbb{R}^n$. Let $i:X\rightarrow 2^\mathcal{M_0}$ be an encoding function. We say that $i\in I_{DAT}$ if, for some $A'\subseteq A$, there exist mappings $g:\mathbb{R}\rightarrow \mathbb{T}$, $o_1,...,o_l:A'\rightarrow 2^{\{1,...,k\}}$, $h_1,...,h_l$ with $h_j:A'\rightarrow \bigcup\limits_{o_j(Q)} X_{[o(_jQ)]}$, and $r_1,...,r_l:A'\rightarrow \{k+1,...,n\}$ with the following properties:
\begin{itemize}
    \item for every $m\in \{k+1,...,n\}$ there exists $j$ and $Q\in A'$ with $r_jQ)=m$
    \item If $Q,Q'\in A'$ and $Q \neq Q'$ then for all $j$ $h_j(Q)\neq h_j(Q')$ or $r_j(Q)\neq r_j(Q')$;
    \item if $M\in i(x)$ and $(x_{o_j(Q)_1},x_{o_j(Q)_2},...,x_{o_j(Q)_l})=h_j(Q)$ then $f(M)(Q)=g(r_j(Q))$.
\end{itemize}  }
\end{definition}
\noindent In the above definition, $o$ associates a set of neurons from the first $k$ neurons to each atom and $X_{[o_l(Q)]}$ represents the subspace of $X$ given by those neurons and $o_l(Q)_i$ represents the $i^{th}$ label from smallest to largest in $o_l(Q)$. For a map in $I_{DAT}$, the first $k$ neurons determine the set of atoms being represented by the state of the network. The truth value of each atom in this set is then given by the remaining neurons. The second condition ensures that although atoms may share the same values for $h_l$ or $r_l$, they can not share the same values for $h_l$ and $r_l$; the pair of them specify a unique atom. The need for possibly many maps $h_1,...,h_l$ comes from the case that atoms might have multiple representations in the network. As we will see, the neurons $\{k+1,...,n\}$ each represent a predicate with specific sequence of terms for input. If a predicate appears multiple times in a knowledge base with different input terms, then a neuron is added for each of these. Depending on the assignment of values to the terms, there could be multiple states representing the same atom. In this case we need multiple pairs, $h_j,r_j$ to capture the multiple possible ways a single atom can be represented in the network. In the following examples only a single function triple $(o,h,r)$ will be required.\\
In Example \ref{excsem}, $x_1$ and $x_2$ define sets of atoms by mapping to a particular grounding of the atoms, and $y_1$ and $y_2$ determine the truth assignment for each atom specified by the state of $x_1$ and $x_2$. To see that this satisfies the definition, let $o(Q)$ be the labels for $x_1$ and $x_2$ for all atoms $Q$, for $z\in \{a,b,c,d\}$ let $h(R_1(z))=h(R_2(z))=(\xi^{-1}(z))$ and $r(R_1(z))$ be the label of $y_1$, $r(R_2(z))$ be the label of $y_2$ and $g$ map $0$ to $False$ and $1$ to $True$. We can see that this satisfies the definition fairly easily. For example, take the state $(x_1,x_2,y_1,y_2)=(1,1,0,0)$ then because $h(R_1(a))=(1,1)=(x_1,x_2)=(x_{o(R_1(a))_1},x_{o(R_1(a))_2})$ and $r(R_1(a))$ maps to the third neuron we have $R_1(a)=g(y_1)=g(0)=False$, similarly we have $R_2(a)=g(y_2)=False$.

Notice that, by setting $k=0$, the second condition ensures injectivity of $r$; because $r$ is also surjective, it is invertible. The definition then reduces to that of $I_{NAT}$ making $I_{NAT}$ a special case of $I_{DAT}$. As we will see in Section \ref{sec:semantic_reg}, this method of encoding has become a popular way to encode first-order logic knowledge bases with a variety of methods including those which can broadly be described as fuzzy differentiable logic operators \cite{diff_fuzz}. To see how this works in practice let us look at an example\\

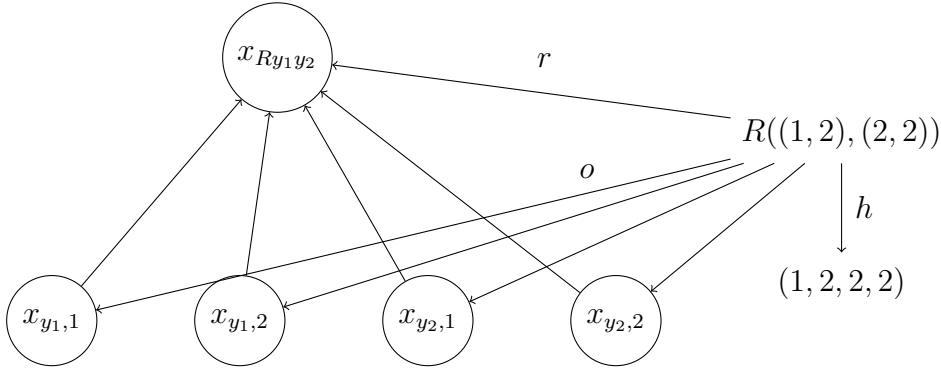
\begin{figure}
\begin{tikzpicture}[scale=1]
\centering
\useasboundingbox (0,0) rectangle (4,4);

 \node[shape=circle,draw=black, anchor=center,minimum size=0.8cm] (a) at (0.5,0.5){$x_{y_1,1}$};

  \node[shape=circle,draw=black, anchor=center,minimum size=0.8cm] (b) at (3,0.5){$x_{y_1,2}$} ;

   \node[shape=circle,draw=black, anchor=center,minimum size=0.8cm] (c) at (5.5,0.5){$x_{y_2,1}$};

  \node[shape=circle,draw=black, anchor=center,minimum size=0.8cm] (d) at (8,0.5){$x_{y_2,2}$} ;

  \node[shape=circle,draw=black, anchor=center,minimum size=0.8cm] (e) at (3.5,4){$x_{Ry_1y_2}$} ;

    \node[anchor=center,minimum size=0.8cm] (f) at (11,3){$R((1,2),(2,2))$} ;

        \node[anchor=center,minimum size=0.8cm] (g) at (11,1){$(1,2,2,2)$} ;

  \coordinate (b_1) at (-0.5,3);
  \coordinate (b_2) at (4.5,3);



 \draw[->] (a) -- (e) ; 
 \draw[->] (b) --  (e) ;
  \draw[->] (c) --  (e) ;
    \draw[->] (d) --  (e) ;

  \draw[->] (f) -- (g) node[near start, below,xshift=0.3cm]{$h$};
  \draw[->] (f) -- (e) node[near end, above,xshift=1.5cm]{$r$};
  \draw[->] (f) -- (a) node[near start, above,xshift=.2cm,yshift=0.1cm]{$o$};
  \draw[->] (f) -- (b) node[near start, above,xshift=.1cm,yshift=-0.1cm]{};
  \draw[->] (f) -- (c) node[near start, above,xshift=0cm,yshift=-0.1cm]{};
  \draw[->] (f) -- (d) node[near start, above,xshift=-0.1cm,yshift=-0.2cm]{};


\end{tikzpicture}
\caption{A figure showing how an atom is represented in a neural network under an encoding in $I_{DAT}$, if $R(y_1,y_2)$ is a term in a first-order language whose interpretations have variable domain $\mathbb{R}^2$, then the atom, $R((1,2),(2,2))$ and its truth value is represented by a state in the network determined by the maps $o,h$, and $r$, if the neurons mapped to by $o$ have values $h(R((1,2),(2,2))$ then the truth value of the atom is given by the neuron $r(R(1,2),(2,2))$}
\label{IDAT_ex}
\end{figure}

\begin{example}
\textit{
Consider a first-order language with two binary predicates $\mathcal{R}=\{R_1,R_2\}$, variables $y_1,y_2,...$, and a function symbol $f$. Assume the domain of the interpretations of this language is $\mathbb{R}^2$. Atoms in the interpretations of this language are of the form $R_i(a_1,a_2)$ where $a_i\in \mathbb{R}^2$. Given the knowledge base $\{\forall y_1,y_2 R_1(y_1,y_2) \Rightarrow R_2(\theta_1,y_2)\}$ where $\theta_1=f(y_1)$, we construct a feed-forward neural network in which the input layer consists of $2\cdot 3=6$ neurons, labeled $x_{y_1,1},x_{y_1,2},x_{y_2,1},x_{y_2,2}$ representing the groundings of $y_1$ and $y_2$, $2$ neurons representing groundings of the term $\theta_1$, labeled $x_{\theta_1,1},x_{\theta_1,2}$, and two neurons representing $R_1(y_1,y_2)$ and $R_2(\theta_1,y_2)$ labeled $x_{R_1y_1,y_2}$ and $x_{R_2\theta_1,y_2}$ respectively. For convenience assume the neurons are given in that order, ie neuron 1 is $x_{y_1,1}$, neuron 2 is $x_{y_1,2}$ etc. Let the transfer function for the final two neurons be the step-function and the transfer function for the other neurons be the identity function.  Let $i$ be the encoding function which maps a state $(x_{y_1,1},x_{y_1,2},x_{y_2,1},x_{y_2,2},x_{\theta_1,1},x_{\theta_1,2},x_{Ry_1,y_2},x_{R\theta_1,y_2}$) to the set of interpretations in which the atoms $R_1(a_1=(x_{y_1,1},x_{y_1,2}),a_2=(x_{y_1,2}),x_{y_2,1})$ and $R_2(a_3=(x_{\theta_1,1},x_{\theta_1,2}),a_4=(x_{y_2,1},x_{y_2,2}))$ are $True$ if their respective neurons have a value of $1$ and $False$ otherwise. We can see this is in $I_{DAT}$ by specifying the functions, $o,h,r$ and $g$. The set of atoms $A'$ is the set of all atoms of the form $R_j(a_1,a_2)$ where $a_1,a_2\in \mathbb{R}^2$ and $j\in \{1,2\}$. $o$ maps atoms with $j=1$ onto $\{1,2,3,4\}$ and atoms with $j=2$ onto $\{3,4,5,6\}$ (ie it associates atoms with the neurons corresponding to the arguments of their predicate), $h$ maps the atom onto the 4 dimensional vector $a_1,a_2$, $r$ maps the atom onto the neuron $x_{Ry_1,y_2}$ if $R_j=R_1$ and $x_{Ry_1,y_2}$ if $R_j=R_2$, and $g$ maps $1$ to $True$ and $0$ to $False$. We check the conditions of the definition, the first is obvious, for the second if $R_j(a_1,a_2) \neq R_j';(a_1',a_2')$ then either $R_j\neq R_j'$ or $(a_1,a_2)\neq (a_1',a_2')$ meaning their values are different for either $r$ or $h$. Finally, given a state of the neural network, if $h(R_j(a_1,a_2))=(x_1,x_2,x_3,x_4)$ for some values $x_1,x_2,x_3,x_4$ which are equal to the values of the neurons in $o(R_j(a_1,a_2))$ in a given state, then, if $R_j=R_1$, by definition $R_j(a_1,a_2)$ has truth value $True$ if the value of $x_{R_1y_1y_2}$ is $1$ and a value of $False$ otherwise. If $R_j=R_2$ then the truth value is determined by $x_{R_2\theta_1,y_2}$. In other words  $R_j(a_1,a_2)=g(r(R_j(a_1,a_2)))$.   }
 \end{example} 
 Note that in the previous example there were no duplicate representations of terms. In other words $R_j(a_1,a_2)$ had a unique representation in the network. In general this will not always be the case as there may be multiple uses of the same predicate in a knowledge base which might mean there are multiple variable assignments that result in the same grounding of the predicate. When this is the case we define additional maps $o_j,h_j,r_j$ for as many duplicate predicates appear in the knowledge base. These maps are identical for each predicate which is not a duplicate but if a predicate is a duplicate then the values of these functions are defined as before but applied to the new predicate-term representation.
 
Common mappings $I_{NAT}$ and $I_{DAT}$ both deal with how atoms should be encoded in a neural network. But there are other conditions that we might require for a set of candidate mappings. When dealing with logical systems that assign real-values to atoms or sentences, we may want to require $i$ to be continuous with respect to appropriate topologies. The relationship between $i$ and the updates of the neurons in $N$ gives additional potential restrictions on the set of candidate mappings. An example of this could be a monotonicity condition. We might say that an encoding function is monotonic if, given an ordering of the truth values, the function satisfies $f(i(x))(q)\leq f(i(N(x))(q)$ for all $x$. There is at least one encoding in the literature with this property \cite{HebbianLogic} but we will not cover it in detail.  It may also be possible to impose more general constraints on sets of candidate mappings by adding structure to logical systems such as Labeled Deductive Systems \cite{LDS}. No matter which set of candidate mappings is being considered, the formal definitions of semantic encoding provided earlier remain the same. The framework should apply to the breadth of logical systems that one may desire to encode semantically in a neural network, with future work looking to define new sets of candidate mappings as the need arises.

\subsection{Probabilistic Encodings}
\label{sec:prob_encod}
So far we have discussed encodings relating deterministic neural networks to logical systems. This leaves out a large number of neural networks and logical systems of relevance to AI, namely, probabilistic models. Luckily, we can generalize the definition of semantic encoding to accommodate these cases. In this sub-section, we cover such a generalization to probabilistic neural networks. As in the deterministic case, we assume that there is a mapping between states of the neural network and interpretations of a logical system as well as an aggregation function. The main difference is that a network now defines a stochastic process with the random variable $X^{(t)}$ representing the state of the network at time $t$. In all examples of probabilistic neural networks that we examine, the state space will be finite. This greatly simplifies the definition of a probabilistic encoding and for this reason we assume that the state space of a probabilistic neural network is finite. We say that $x$ is a \emph{limit point} if there exists $\epsilon>0$ such that for all $t>0$, there exists $t'>t$ with $P(X^{(t')}=x)>\epsilon$. Call the set of such points $X_{P,inf}$. 
\begin{definition}
\textit{Let $N$ be a neural network with state space $X$ and a corresponding stochastic process $\{X^{(t)}\}_{t=0}^{\infty}$. Let $\mathcal{S}=(\mathcal{L},\mathcal{M})$ be a logical system and $L$ a knowledge-base of $\mathcal{S}$. Given a mapping $i:X\rightarrow 2^{\mathcal{M}}$ and an aggregation function $Agg:2^{2^{\mathcal{M}}}\rightarrow 2^{\mathcal{M}}$, we define the following:
\begin{itemize}
    \item  Let $\mathcal{M}_N=Agg(\{i(x)| x\in X_{P,inf}\})$. $N$ is a probabilistic neural model of $L$ under $I$ and $Agg$ if $\emptyset \subset \mathcal{M}_N\subseteq \mathcal{M}_L$.
    \item  $N$ is a probabilistic semantic encoding of $L$ under $I$ and $Agg$ if $N$ is a probabilistic neural model of $L$ under $i\in I$ and $Agg$, and $\mathcal{M}_N\subseteq L'$ iff $L\vDash_{\mathcal{S}}L'$. 
\end{itemize}
}
\end{definition}
In a probabilistic neural model, the probability that the state of the neural network represents information about models of $L$ converges to $1$. We know this because, for a finite state-space, $\lim\limits_{t\rightarrow \infty} P(X^{(t)}\not\in X_{P,inf})=0$. For neural networks without a finite state space, this definition, along with that of $X_{P,inf}$ would have to be generalized as it could be the case that $P(X^{(t)}\not\in X_{P,inf})>0$ for all $t>0$.  

Notice that we can define a stochastic process that is equivalent to the underlying deterministic network by setting $P(X^{(t)}=N^t(x_0)|X^{(0)}=x_0)=1$, with $P(X^{(0)})$ being the uniform distribution. In this process, the state of the network updates according to $N$ with probability $1$ and all other possible states have probability $0$. Consider the set of limit points, $X_{P,inf}$, in this process. These are the set of points, $x$, such that there exists a sequence of time points $t_1',t_2',...$ with the property that there exists $t_i'$ with $P(X^{(t_i')}=x)>\epsilon$ for some $\epsilon$, but this is only true if there exists $x_0\in X$ with $N^{t_i'}(x_0)=x$. This means that $x$ is infinitely recurring. Conversely, if $x$ is infinitely recurring then there exists a sequence $t_1',t_2',...$ and some $x_0$ such that $N^{(t_i')}(x_0)=x$, that is, $x\in X_{P,inf}$. This means that $X_{\textit{inf}}=X_{P,\textit{inf}}$ and that $N$ is a probabilistic semantic encoding under $I$ and $Agg$ if and only if it is a semantic encoding according to Definition \ref{semanticneuralmodel}. This shows that, when the state-space is finite, the probabilistic definition is a generalization of the deterministic definition of semantic encoding. 

The stochastic property corresponding to the deterministic property of stability is the existence of a stationary distribution. Let $\hat{P}$ be a distribution on $X$. We say that $\hat{P}$ is stationary with respect to the stochastic process if whenever the initial distribution, $X^{(0)}$, is equal to $\hat{P}$ then the distribution $X^{(t)}$ is also equal to $\hat{P}$ for all $t>0$. In some cases, a stationary distribution is also a limiting distribution, that is, $\lim\limits_{t\rightarrow \infty} P(X^{(t)}=x)=\hat{P}(x)$. In the case that we have a single stationary distribution which is also a limiting distribution, $X_{P,inf}=\{x|x\in X, \hat{P}(x)>0)\}$, it becomes easy to prove that a network is a semantic encoding by examining the states with non-zero probability in the limiting distribution (as done in \cite{MLN} for example).

\subsection{Approximate Encodings}
\label{sec:approx_encode}
The notion of semantic encoding defines when a neural network \emph{implements} a knowledge-base (as if the knowledge-base were the specification of the network). This is useful when we know that the data must satisfy some prior constraints. However, in practice, requiring background knowledge to be a hard constraint can be too restrictive. In many cases, observed data may contradict our background knowledge, in which case we must decide whether to trust more our knowledge about the world or our observations of the world. To this end, many recent neuro-symbolic methods have taken an approximate approach to neural encoding. Rather than selecting architecture and weights that directly encode a knowledge-base, such methods define a function that measures how far a neural network is from being a semantic encoding of a given knowledge-base. This function is then added to the loss function of the network as a regularization. We will explore these methods in more detail in Section \ref{sec:semantic_reg} but for now, we introduce the notion of a fidelity measure, upon which these methods rely. 
A fidelity measure is a function that measures how far a neural network with given encoding and aggregation functions is from being a neural model of a particular knowledge-base. As in the previous sections, we refrain from imposing hard requirements on the definition of a fidelity measure as there are different measures used in the literature and we do not wish to make choices at this point regarding exactly what properties a fidelity measure should have. With this in mind, we define a fidelity measure as follows.
\begin{definition}
\textit{Let $\mathcal{N}$ be a set of neural networks and $\mathcal{S}=(\mathcal{L},\mathcal{M})$ a logical system. Define the set $\mathcal{N}\times {I}\times AGG$ as the set of all triples $(N,i,Agg)$, where $i:X\rightarrow 2^{\mathcal{M}}, i\in I$, $X$ is the state space of $N$, and $Agg:2^{2^{\mathcal{M}}}\rightarrow 2^{\mathcal{M}}$. We say that $Fid: (\mathcal{N}\times I \times AGG) \times 2^{\mathcal{L}}\rightarrow [0,1]$ is a \textit{fidelity measure} when $Fid((N,i,Agg),L)=1$ if and only if $N$ is a neural model of $L$ under $i$ and $Agg$.} 
\end{definition}
Despite the general definition, the examples that we will examine in Section \ref{sec:semantic_reg} will all use one of two ways of measuring fidelity. We introduce these two ways next. 

Logic Tensor Networks and many similar methods encode knowledge-bases using variants of first-order fuzzy logic. When this is the case, each interpretation and formula pair is assigned a value in $[0,1]$ representing the degree to which the interpretation satisfies the formula. This can be used to define a fidelity measure, as follows.

Let $\mathcal{S}=(\mathcal{L},\mathcal{M})$ be a fuzzy logic where $\mathcal{L}$ consists of sentences of the form $[a,b] : \phi$ where $\phi$  is a sentence of propositional or first-order logic and $0\leq a \leq b \leq 1$ (we can recover the unlabelled case by setting $a=b=1$ for every sentence). Each $M\in \mathcal{M}$ defines a function $M(\phi)\in [0,1]$ where $\phi$ is the unlabelled part of a sentence in $\mathcal{L}$. $M$ is a model of $L$ iff for all $[a,b]:\phi \in L$, $M(\phi)\in [a,b]$. The function $M$ is generally defined by interpreting logical symbols and quantifiers as fuzzy connectives but the following definition will work for any system in which every interpretation has a corresponding function of this form. 
\begin{definition}
Define $Fid_{fuzzy}:(\mathcal{N}\times I \times AGG) \times 2^{\mathcal{L}}\rightarrow [0,\infty)$ by $Fid_{fuzzy}((N,i,Agg),L)=\inf\limits_{M\in \mathcal{M}_N} \SatAgg\limits_{[a,b]:\phi \in L} \{1-d(M(\phi),[a,b])\}$
\end{definition}
$\mathcal{M}_N$ is defined via $N$, $i$ and $Agg$, $\SatAgg$ is a function aggregating the satisfiability scores for each sentence in $L$, and $d(M(\phi),[a,b]))$ is the distance from the real number $M(\phi)$ to the interval $[a,b]$. In most practical cases, $L$ is finite, in which case $\SatAgg$ can be chosen to be the mean or the minimum, but some papers keep the formulation general. All that is required for this to be a fidelity measure is that $\SatAgg\{x_1,x_2,x_3,...\}\leq 1$ and also that $\SatAgg\{x_1,x_2,x_3,...\}=1$ if and only if $x_1=x_2=x_3=...=1$. 
\begin{lemma}
If  $\SatAgg\{x_1,x_2,x_3,...\}\leq 1$ and also $\SatAgg\{x_1,x_2,x_3,...\}=1$ if and only if $x_1=x_2=x_3=...=1$ then $Fid_{fuzzy}$ is a fidelity measure.
\end{lemma}
\begin{proof}
    
To see this, consider $Fid_{fuzzy}$ with $\SatAgg$ satisfying the stated conditions. First note that $\SatAgg\leq 1$ means that $Fid_{fuzzy}((N,i,Agg),L)\in [0,1]$. Now assume that $N$ is a neural model of $L$ under $i$ and $Agg$, that is, for all $M\in \mathcal{M}_N$ and $[a,b]: \phi \in L$, $M(\phi) \in [a,b]$ and thus $d(M(\phi),[a,b])=0$ meaning for all $M \in \mathcal{M}$,
$\SatAgg\limits_{[a,b]:\phi \in L} \{1-d(M(\phi),[a,b])\}=\SatAgg(\{1-0,1-0,1-0,...\})$ so
$Fid_{fuzzy}((N,i,Agg),L)=1$.

Conversely, assume $Fid_{fuzzy}((N,i,Agg),L)=1$, that is, $\inf\limits_{M\in \mathcal{M}_N} \SatAgg\limits_{[a,b]:\phi \in L} \{1-d(M(\phi),[a,b]\}=1$. Because $\SatAgg \leq 1$, the only way this can be true is if $\SatAgg\limits_{[a,b]:\phi \in L} \{1-d(M(\phi),[a,b]\}=1$ for all $M\in \mathcal{M}_N$, but by assumption this means that for all $M$ and all $[a,b] : \phi \in L$, $d(M(\phi,),[a,b])=0$, which implies that $M$ is a model of $L$. Hence, $ \mathcal{M}_N \subset \mathcal{M}_L$ and $N$ is a neural model of $L$ under $i$ and $Agg$.  
\end{proof}
We illustrate the use of this fidelity measure with an example.
\begin{example}
\textit{
Recall Example \ref{exmp:1} in which we defined a simple recurrent neural network to be a semantic encoding of the knowledge-base $\{(A\wedge B) \vee (\neg A \wedge \neg B)\} $. Consider the fuzzy knowledge-base $L= \{ [0.75,1] : A \vee B, [0.5,1] : \neg A \vee \neg B\}$. Using the same aggregation and encoding function as in Example \ref{exmp:1} and $\SatAgg=\min$, we have that $\mathcal{M}_N= \{M | M(A)=M(B)=1\} \cup \{M |M(A)=M(B)=0\}$. In the first case, $M(A \vee B)=1$ and $M(\neg A \vee \neg B)=0$ so $\SatAgg\limits_{[a,b] : \phi \in L} \{1-d(M(\phi),[a,b])\}= \SatAgg \{1-d(1,[0.75,1]),1-d(0,[0.5,1])\}=\SatAgg\{1,0.5\}=0.5$. In the second case, $M(A \vee B)=0$ and $M(\neg A \vee \neg B)=1$ so $\SatAgg\limits_{[a,b] : \phi \in L} \{1-d(M(\phi),[a,b])\}= \SatAgg \{1-d(0,[0.75,1]),1-d(1,[0.5,1])\}=\SatAgg\{0.25,1\}=0.25$. Thus, for each $M\in \mathcal{M}_N$, $\SatAgg$ is either $0.25$ or $0.5$ meaning that $Fid_{fuzzy}((N,i,\cup),L)=\inf\{0.5,0.5,...,0.25,0.25,...\}=0.25$}.
\end{example}
The other main fidelity measure used is a probabilistic fidelity measure. This measure looks at the \textit{probability} that the network is in a state that represents a model of the knowledge-base. Note that this measure is only appropriate for encodings using $Agg=\cup$. When this is the case, $\mathcal{M}_N=\cup_{x\in X_{P,inf}} i(x)$ and we can check whether or not $\mathcal{M}_N \subset \mathcal{M}_L$ by checking whether $i(x)\subseteq \mathcal{M}_L$ for all $x\in X_{P,inf}$.
\begin{definition}
Given a set of probabilistic neural networks, $\mathcal{N}$, each with a finite state space and a unique limiting distribution, define $Fid_{prob}:(\mathcal{N}\times \mathcal{I} \times AGG) \times 2^{\mathcal{L}}$, where $AGG=\{\cup\}$, to be $Fid_{prob}:((N,i,\cup),L)=\hat{P}(i(x)\subseteq \mathcal{M}_L)$, where $\hat{P}$ is the limiting distribution of $N$.
\end{definition}
\begin{lemma}
    $Fid_{prob}$ as defined above is a fidelity measure.
\end{lemma}
\begin{proof}
    
To see that this is a fidelity measure, first note that its value is always contained in $[0,1]$. Next assume $(N,i,\cup)$ is a neural model of $L$, then $\cup_{x \in X_{P,inf}} i(x) =\mathcal{M}_N\subseteq \mathcal{M}_L$ so $i(x)\subseteq \mathcal{M}_L$ for all $x\in X_{P,inf}$. Because $N$ has a unique limiting distribution, as discussed in Section \ref{sec:prob_encod}, $X_{P,inf}=\{x|x\in X, \hat{P}(x)>0\}$. Thus, by definition $\hat{P}(x\in X_{P,inf})=1$, and $\hat{P}(i(x)\subseteq \mathcal{M}_L)=1$. Conversely, if $\hat{P}(i(x)\subseteq \mathcal{M}_L)=1$, then because $x\in X_{P,inf}$ iff $\hat{P}(x)>0$, we have that $x\in X_{P,inf}$ if and only if $i(x)\subseteq \mathcal{M}_L$. Thus, $\mathcal{M}_N=\cup_{x \in X_{P,inf}} i(x) \subseteq \mathcal{M}_L$ making this a fidelity measure.
\end{proof}
In line with the methods that we will describe in Section \ref{sec:semantic_reg}, we will illustrate $Fid_{prob}$ with an example using a feed-forward network in which the final layer is stochastically determined by the previous layers as follows. Note that such a network settles to a unique stationary distribution given an initial distribution over the input layer
\begin{example}
    Consider a feed-forward network, $N$, with a single binary input, $x$ and two binary output neurons, $y_1,y_2$. Let $N$ define a stochastic process over the neurons $x,y_2,y_2$ in the following way, $y_1$ and $y_2$ are conditionally independent given $x$, $P(y_1^{(t+1)}=1 |x^{(t)}=1)=0.4$, $P(y_2^{(t+1)}=1 | x^{(t)}=1)=0.3$, $P(y_1^{(t+1)}=1 | x^{(t)}=0)=1$, $P(y_2^{(t+1)}=1 | x^{(t)}=0)=0.2$, and $P(x^{(t)}=1)=0.5$ for all $t\geq 0$. Consider an encoding function $i\in I_{NAT}$, mapping the neurons $x,y_1,y_2$ to propositional variables $X,Y_1,Y_2$ in propositional logic. Consider the knowledge-base $\{Y_1\vee Y_2, \neg (Y_1 \wedge Y_2)\}$. The states of the network which satisfy this knowledge-base are those with $y_1=1,y_2=0$ or $y_1=0,y_2=1$. The probability of these states in the limiting distribution is $\hat{P}(y_1=y_2=1)=(0.4\cdot 0.3)\cdot 0.5 + (1 \cdot 0.2)\cdot 0.5 = 0.16$, $\hat{P}(y_1=0,y_2=0)=(0.6\cdot 0.7)\cdot 0.5 + (0\cdot 0.8)\cdot 0.5= 0.21$, making the total probability for the network to be in one of these states $0.37$, i.e. $Fid_{prob}((N,i,\cup),L)=0.37$.
\end{example}
There are many other possible choices for fidelity measure. A generic choice, that can be used for any logical system in which there is a metric defined on the set of interpretations, is $d(\mathcal{M}_n,\mathcal{M}_L)$ where $d(\cdot,\cdot)$ is the Hausdorff distance between sets. In most of the examples in the literature, however, $Fid_{fuzzy}$ and $Fid_{prob}$ are the ones used.

\subsection{Towards a Theory of Neuro-symbolic Computation}
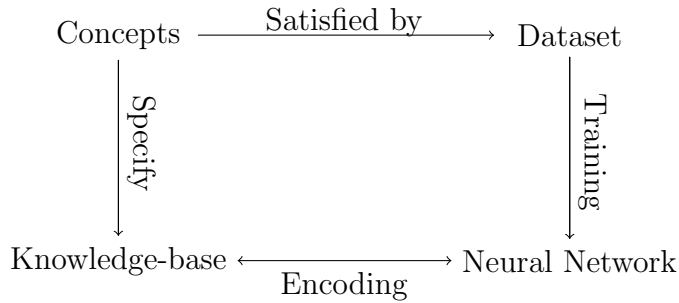
\begin{figure}
\centering
\begin{tikzpicture}

\useasboundingbox (0,0) rectangle (4,4);

 \node[shape=rectangle,draw=none, anchor=center,minimum width=2.1cm] (x_1) at (-1.5,3.5){Concepts} ;

  \node[shape=rectangle,draw=none, anchor=center, minimum width=2cm] (x_2) at (-1.5,0.5){Knowledge-base};
  
  \node[shape=rectangle,draw=none, anchor=center, minimum width=2cm] (y_1) at (4.5,3.5){Dataset};
  
  \node[shape=rectangle,draw=none, anchor=center, minimum width=2cm] (y_2) at (4.5,0.5){Neural Network};
\path[draw=black,
      postaction={decorate},
      decoration={text along path,
                  text=Specify,
                  raise=0.2cm,
                  text align={left indent={0.2\dimexpr\pgfdecoratedpathlength\relax}}
                  }
                  ] 
(x_1) -- (x_2);

\path[draw=black,
      postaction={decorate},
      decoration={text along path,
                  text=Training,
                  raise=0.2cm,
                  text align={left indent={0.2\dimexpr\pgfdecoratedpathlength\relax}
                  }
                  }
                  ] 
(y_1) -- (y_2);

 \draw[->] (x_1) -- (y_1) node[ below,yshift = 0.5cm,xshift=-3cm]{Satisfied by};
  \draw[<->] (x_2) --  (y_2)  node[ below,yshift = -0.01cm,xshift=-3cm]{Encoding};
  
  \draw[->] (y_1) -- (y_2);
  \draw[->] (x_1) -- (x_2);

\end{tikzpicture}
\caption{Relationship between a neural network trained on a data set and a knowledge-base representing relations among the concepts of a given task.}
\label{hypot}

\end{figure}

In neuro-symbolic AI, domain knowledge is either learned from data and made explicit with the use of explainable AI methods, or it is available in the form of partial background knowledge about the underlying task that can be revised from data. Allowing any such explicit knowledge to benefit the training of a neural network is the main goal of an encoding. Standard neural networks can only learn from data. Neuro-symbolic networks learn from data and knowledge. An adequate encoding is therefore expected to make it easier for the network to learn the task correctly, either by making it faster, use fewer resources, or to make the network satisfy certain constraints that are not necessarily present in the data. Consider the goal of making learning converge faster by enabling the use of background knowledge alongside training from data. Here, the goal is to reduce the search-space by excluding possible worlds which are known \textit{a priori} to violate a certain property of the task at hand. The majority of the efforts in this area have adopted a practical approach: taking various methods of neural encoding and comparing their performance to techniques that do not encode any prior knowledge. Much less attention has been paid to theoretical results that, we argue, should guide the development of effective encoding, such as the provision of soundness proofs, as advocated in \cite{CILP}. A theory of effective encoding would allow the field to start asking fundamental questions such as: What properties does the learning algorithm need to have to guarantee effective encoding? Does adding background knowledge improve the generalization ability of networks on every dataset or is it limited in certain cases? These questions have come into sharper focus with recent results about \textit{reasoning shortcuts} in neural networks \cite{marconato2023neurosymbolic}, which show that certain encoding methods will inevitably, under some circumstances, achieve high accuracy by leveraging concepts in the wrong way. That is to say, abstract concepts from background knowledge can be assigned an unintended meaning and still improve training performance. An unsound encoding, therefore, may exacerbate the well-known effect of neural network training that provides \textit{the right answer for the wrong reason}. Hence, semantics matter. If concepts are learned by association with confounding factors, it compromises generalizability and interpretability. This highlights the need for a much more robust general theory of neural encoding to identify the properties required for neural networks to learn how to correctly reason about a dataset. The development of such a theory is hamstrung by a lack of standardization in the field. The fact that different encoding methods use different definitions means results can only be developed for very narrow families of neural encoding, despite the likelihood that similar results apply to a whole host of encoding methods. Figure \ref{hypot} illustrates the relationships at play that will need to be formalized by a general theory.
\\\\

\noindent We argue that the framework presented here has the potential to be the foundation for a unified theory of neuro-symbolic computation. Two things need to be true for this to be the case. First, the definitions must capture a sufficient number of the encodings used in the literature. If none of the encodings used in practice are semantic encodings according to our definition, then any theory developed for semantic encodings will not apply to the existing techniques in the field and will have no utility. In the following sections we show that a substantial portion of the existing encoding techniques satisfy the definition of semantic encoding; this is the primary claim of this paper. The second thing that must be true is that the definitions must be robust enough to generate meaningful insights into the properties of the methods that they describe. This will only become clear in the course of developing a theory, something which is out of the scope of a single paper. In fact, it may be the case that future research reveals alternative definitions and frameworks that are better able to produce theoretical insight. In this light, our framework can be seen as a jumping-off point from which the beginnings of a theory can take shape. We will, however, outline some properties of semantic encoding that make our definitions promising candidates for a theoretical foundation. To start with, we note the following transitive properties of encodings.

\begin{theorem}
\label{theorem:trans_1}
\textit{Given two neural networks, $N_1, N_2$, and a bijection, $f$, between their state spaces, $X,Y$, satisfying $f(N_1(x))=N_2(f(x))$, then if $N_2$ is a semantic encoding of a knowledge-base $L$ in a logical system, $\mathcal{S}$, under $i\in I$ and $i\circ f \in I$ then $N_1$ is a semantic encoding of $L$ under $i\circ f$}.
\end{theorem}

\begin{theorem}
\label{theorem:trans_2}
    Let $\mathcal{S}_1=(\mathcal{M}_1,\mathcal{L}_1)$ and $\mathcal{S}_2=(\mathcal{M}_2,\mathcal{L}_2)$ be logical systems with maps $g:\mathcal{M}_2\rightarrow\mathcal{M}_1$, and $f:\mathcal{L}_2\rightarrow\mathcal{L}_1$ with the following properties:
    \begin{itemize}
        \item $g$ is bijective
        \item $m\in \mathcal{M}_2$ is a model of $l \in \mathcal{L}_2$ iff $g(m)$ is a model of $f(l)$
        \item for $L_1,L_2\in \mathcal{L}_2$, $L_1\vDash_{\mathcal{S}_2} L_2$  iff $f(L_1) \vDash_{\mathcal{S}_1} f(L_2)$
    \end{itemize}
    then if a network $N$ is a semantic encoding of $f(L)$ under $i\in I$ and $g^{-1} \circ i \in I$ then $N$ is a semantic encoding of $L$ under $g^{-1} \circ i \in I$.
\end{theorem}
\begin{cor}
if the conditions of Theorem \ref{theorem:trans_2} hold for $\mathcal{S}_1$ and $\mathcal{S}_2$, and $\mathcal{S}_1$ is semantically equivalent to a set of networks $\mathcal{N}$ under $I$ and $g^{-1} \circ i \in I$ for each encoding function, $i$, then there exists a subset $\mathcal{N}'\subset \mathcal{N}$ that is semantically equivalent to $\mathcal{S}_2$ under $I$.
\end{cor}

The proof of each of these is relatively straightforward and can be found in Appendix A. This solves the problem of when a neural encoding of a logical system can be extended to other, equivalent logical systems. As an example, in the next section  we will look at the encoding of Penalty Logic into Hopfield networks. Propositional Logic can be expressed in Penalty Logic in such a way that the conditions of Theorem \ref{theorem:trans_2} are satisfied. This means that we know that the encoding for Penalty Logic can also be used as an encoding for Propositional Logic. 
\\\\
The other property we wish to highlight is that the standard supervised learning classification task in Machine Learning can itself be framed as a type of semantic encoding. Given the task of training a neural network classifier, i.e. to learn a mapping $f:X_{input}\rightarrow Y$, where $X_{input}\subset \mathbb{R}^n$ is the set of inputs and $Y=\{1,2,3,...,n\}$ is a set of labels, we can translate the problem into the equivalent problem of training a neural network to be a semantic encoding of a specific knowledge-base. To do this we create a first-order language by defining a predicate for each label, i.e. for each $j\in y$ there is a predicate $R_j$ in $\mathcal{L}$. Given a finite training and test set, $X_{train},X_{test}\subset X_{input}$, define a constant symbol, $c_x$, for each $x\in X_{train}\cup X_{test}$. Now, we define a set of first-order interpretations for this language by setting the domain to be $X_{input}$ and requiring each constant $c_x$ to be mapped to the corresponding input, $x\in X_{train}\cup X_{test}$. With this simple logical system, the training set can be defined as a knowledge-base, $L_{train}$ consisting of all statements of the form $R_m(c_x)$ where $x\in X_{train}$ and $f(x)=m$. Likewise, we define $L_{test}$. A standard feed-forward neural network trained on this dataset can now be seen as encoding interpretations of this language with an encoding function under $I_{DAT}$ and $Agg=\cap$. Furthermore, measuring the accuracy of the network satisfies our definition of a fidelity function. The process of training a neural network is then equivalent to maximizing $Fid((N,i,\cap),L_{train})$ with the goal of maximizing $Fid((N,i,\cap),L_{test})$. By translating the standard learning process into the language of semantic encoding, we should be able to start asking general questions about different semantic encoding techniques, such as which encoding of knowledge-base $L$ results in the highest fidelity, thus relating the different learning algorithms with a formal semantics.

This example shows how a theory of semantic encodings could act as a bridge between neuro-symbolic methods and the development and analysis of learning algorithms. Take some learning algorithm $\mathcal{A}$, such as gradient descent, this algorithm uses the loss on the training set to update the parameters of the network with the goal of minimizing the loss on the test set. Carrying on from our previous formulation, this can be expressed as $\mathcal{A}$ being a mapping with $\mathcal{A}(Fid((N,i,\cap),L_{train}), N) = N'$ where $N'$ is the network with updated weights. The goal then is to maximize $Fid((\mathcal{A}^k(N_0),i,\cap),L_{test})$ where $ \mathcal{A}^k(N_0)$ is the result of applying algorithm $\mathcal{A}$ $k$ times to an initial network $N_0$. The question of how beneficial background knowledge, $L$, will be can then be seen as looking at the effect of applying $\mathcal{A}$ to $N$ with fidelity term $Fid((N,i,\cap),L_{train}\cap L)$. Or alternatively if $\mathcal{N}$ is a set of networks that are all models of $L_{background}$ then does restricting $\mathcal{A}$ to $\mathcal{N}$ give better results\footnote{The former technique is equivalent to adding a soft constraint on the loss function while the latter is equivalent to fixing the architecture of the network so that it must satisfy the constraints regardless of its weights}? This gives several avenues of investigation, for starters, what properties does $L$ need to have in order for this to be beneficial? If the number of models of $L_{train}\cap L$ is fewer than that of $L_{train}$ then will $\mathcal{A}$ be more likely to produce a better result for $L_{test}$? We can use these questions to examine properties of $\mathcal{A}$ that may be necessary for this to be true. For example, if $x\in X_{train}$ satisfies background knowledge $L$, then we would hope that $\mathcal{A}$ trained on $x$ could not decrease the fidelity of $N$ with respect to $L$. 

How much insight can be gained from this line of inquiry is the task of future work. For the time being, in the next section, we focus on showing that many, if not most, existing encoding techniques satisfy our definition of a semantic encoding.

\section{Many Neuro-symbolic Approaches are Semantic Encodings}
\label{survey}
In this section, we demonstrate that a large number of neuro-symbolic approaches, old and new, satisfy our definition of a semantic encoding. Showing this requires: (a) identifying the encoding function used by each approach and, if necessary, showing that the encoding belongs to the set of candidate maps from Section 3; (b) defining the aggregation function, and (c) showing that under these functions the neuro-symbolic approach specifies a semantic encoding. This may become intuitively clear by illustrating the state transition diagram that corresponds to the neural network and identifying the stable states with their corresponding sets of interpretations (as done in Figures \ref{nnfig:3} and \ref{nnfig:4}). However, the details can become convoluted. With the vast number of encoding techniques, going through each example and showing that they are a semantic encoding is not feasible. Instead, we shall look at larger families of related encoding techniques and provide theorems which can be used to quickly identify any encoding technique in the family as a semantic encoding. We also provide detailed proofs for neuro-symbolic approaches that can be seen as representative of a large class of encoding techniques, such as LTN. The proofs are provided in the Appendix for those who are interested in the technical details of relating their encoding to a semantic encoding. Our main goal with this section is to provide evidence that a large number of methods in the literature satisfy our semantic framework's definitions. We do not claim that our list is exhaustive. At the end of the section, we briefly overview some encoding techniques that were left out, some of which do not fit into our framework. 
\subsection{Logic Programming Semantics}
\label{survey:LP}
We begin our analysis with neuro-symbolic systems based on logic programming. Logic programming has played a central role in the development of neuro-symbolic computing. One of the oldest neuro-symbolic reasoning techniques has been a semantic encoding of logic programs \cite{original1}. Because of this, semantic encodings of various kinds of logic programming exist today \cite{NeSybook1}, with the majority of them making use of the same basic technique. This will allow us to broadly cover various logic programming approaches and extensions. 

We start by considering logic programs using our definition of a logical system. Let a logic programming language contain sentences which consist of \textit{clauses} of the form $A\leftarrow B_1\wedge B_2 \wedge... \wedge B_n$, where $A$ is called the \textit{head} of the clause and $B_1\wedge B_2 \wedge... \wedge B_n$ is the (possibly empty) \textit{body}.\footnote{The body of a logic program is often written as $B_1,B_2,...,B_n$ with the use of commas to denote logical conjunction. Instead, we use $\wedge$ to denote conjunction in order to avoid confusion with the collection of literals in the body of a clause.} The components of the head and body are \textit{literals} which are either ground atoms or their negation. When we refer to first-order logic programming then every variable appearing in each literal of a clause is assumed to be universally quantified. A clause with variables is therefore considered shorthand for the collection of clauses resulting from all groundings of that clause with constants of the language. For example, if the constants of the language are $c$ and $d$ then the clause $A(x)\leftarrow B_1(x)$, meaning $\forall x (A(x)\leftarrow B_1(x))$, represents the clauses $A(c)\leftarrow B_1(c)$ and $A(d)\leftarrow B_1(d)$. A knowledge-base in logic programming, referred to as a logic program, is a set of clauses. A clause in which none of the literals is negated is called a \textit{Horn Clause}. The interpretations of a logic program are truth-assignments to the ground atoms. For our purposes, the set of truth-values will be $\mathbb{T}=\{true,false\}$, although systems of logic programming with many-valued and probabilistic semantics exist \cite{raedt}. A model, $M$, of a logic program, $P$, is a truth-assignment that satisfies $M(A)=1$ if and only if there exists a clause in $P$ with $A$ as head and $M(B_i)=1$ for each $B_i$ in the body of the clause. It is customary to represent interpretations of logic programming as sets containing all atoms which are \emph{true} in the interpretation. So, rather than writing $M(A)=1$ as we have been, we write $A\in M$.

The basic method of semantic encoding for logic programming is known as the CORE method \cite{CORE,CILP}. The CORE method is a way of creating a semantic encoding by using a neural network to implement the \textit{least fixed point operator} of a logic program. The least fixed point operator, $T_P$, of a logic program, $P$, is a mapping from interpretations to interpretations of the language of $P$ defined as $T_P(M)=\{A | \exists (A\leftarrow X_1\wedge X_2 \wedge...\wedge X_k) \in P, X_1,X_2,...,X_k \in M\} $. In other words, $A$ is \emph{true}, $A \in T_P(M)$, if and only if there exists a clause in $P$ with $A$ in the head and with all literals in the body being \emph{true}, i.e. belonging to $M$. What makes $T_P$ useful is that for many classes of logic programs, $T_P$ will always converge to a fixed point that is minimal with respect to the partial ordering of interpretations defined by $M\leq M'$ iff $A\in M \Rightarrow A \in M'$. Furthermore, in these cases, the models of $P$ are exactly these fixed points \cite{fixedpointlogic}. The CORE method implements logical reasoning in neural networks by translating this fixed point operator into the network. The basic methodology is to represent each atom that appears in $P$ as a neuron (although other, more complicated representations have been used) such that for each clause $(A\leftarrow X_1\wedge X_2 \wedge ... \wedge X_k) \in P$, a connection is added from each neuron representing $X_1,X_2,...,X_k$ to the neuron representing $A$ in such a way that if a network state, $x$, maps $X_1\wedge X_2 \wedge ... \wedge X_k$ to \emph{true} then $N(x)$ will map $A$ to \emph{true}. We will elaborate on the details of this procedure in the examples to follow, but for now we give the formalization of the CORE method using our framework of semantic encoding. 

First, we note a slight abuse of notation: an encoding function, $i \in I$, is defined as a mapping to a set of interpretations, but in the examples to follow, $i$ will map to a single interpretation as all atoms that do not appear in the computation will be assigned a truth-value $false$. Because we are assuming that $i$ maps to a unique interpretation, for convenience of notation, we treat $i$ as a mapping $i:X\rightarrow \mathcal{M}$ rather than $i:X\rightarrow 2^{\mathcal{M}}$. Thus, if we write, for example, $A\in i(x)$, we mean that $A$ is \emph{true} in the unique interpretation mapped to by $i$. The theorem below still holds if we allow $i$ to map to sets of interpretations, but in all examples derived from the CORE method, $i(x)$ will consist of a single interpretation and the assumption simplifies the proof.
\begin{theorem}
\label{lemma:tp}
\textit{
Let $\mathcal{S}$ be a logic programming system. Let $P$ be a logic program of $\mathcal{S}$ with models $\mathcal{M}_P$, $T_P$ be the least-fixed point operator of $P$, and $N$ be a neural network with state space $X$ and an encoding $i\in I$, with $i(x)$ mapping to a single interpretation. $N$ is a semantic encoding of $P$ under $I$ with $Agg=\cup$ if the following hold:
\begin{itemize}
    \item $\mathcal{M}_P\subseteq range(i)$;
    \item $T_P$ and $N$ always converge to a fixed-point; and
    \item $i(N(x))=T_P(i(x))$. 
\end{itemize} }
\end{theorem}
\begin{proof}
We know from logic programming that the models of $P$ are exactly the fixed points of $T_P$ \cite{fixedpointlogic}. Thus, if $M$ is a model of $P$ then $T_P(M)=M$. By assumption, $\mathcal{M}_P\subseteq range(i)$ so $M= i(x)$ for some neural configuration $x$. We have that $i(x)=M=T_P(M)=T_P(i(x))=i(N(x))$. By assumption, there is $t>0$ such that $N^{t}(x)$ is a fixed point of the network. Thus, $N^{t}(x)\in X_{inf}$ and applying the previous identity $t$ times yields $i(N^{t}(x))=T_P^t(i(x))=M$. Because $Agg=\cup$, we have that $M\in \mathcal{M}_N$. Conversely, if $M\in \mathcal{M}_N$, because $Agg=\cup$, $M=i(x)$ for some fixed point $x$. Furthermore, $i(x)$ must be a stationary point of $T_P$ because $i(x)=i(N(x))=T_P(i(x))$. Therefore, $M\in \mathcal{M}_P$. This gives us that $\mathcal{M}_N=\mathcal{M}_P$, making $N$ a semantic encoding of $P$ under $I$ and $Agg=\cup$.
\end{proof}
\noindent We can now survey some of the rule-based encodings and, with the help of the previous theorem, show how they fit into our framework for semantic encoding. We begin our survey with neuro-symbolic learning system KBANN \cite{KBANN} for which, differently from the CORE method, no proof of soundness existed up to now. We consider the simplified case without negation where KBANN encodes acyclic Horn clauses into binarized feed-forward neural networks.\footnote{A logic program is acyclic when there are no cycles through the atoms in the heads of the clauses. This can be formally defined by using a logic program to construct a graph such that each atom in the logic program is a node in the graph and there is a connection from node $A$ to node $B$ if there is a clause in the logic program with $A$ in the body and $B$ in the head. If the graph contains no cycles then the logic program is said to be acyclic.} Differently from the CORE method, KBANN is end-to-end differentiable, using a sigmoid activation function instead of a step function. With our framework, we will provide a semantic equivalence between acyclic Horn clauses and binarized feed-forward networks with positive weights under $I_{NAT}$.
\begin{proposition}
\textit{
Binarized feed-forward networks with positive weights are semantically equivalent to acyclic Horn clauses under $I_{NAT}$.}
\end{proposition}
\begin{proof}
See Appendix B.
\end{proof}

\begin{figure}
    \centering
    \includegraphics[scale=0.5]{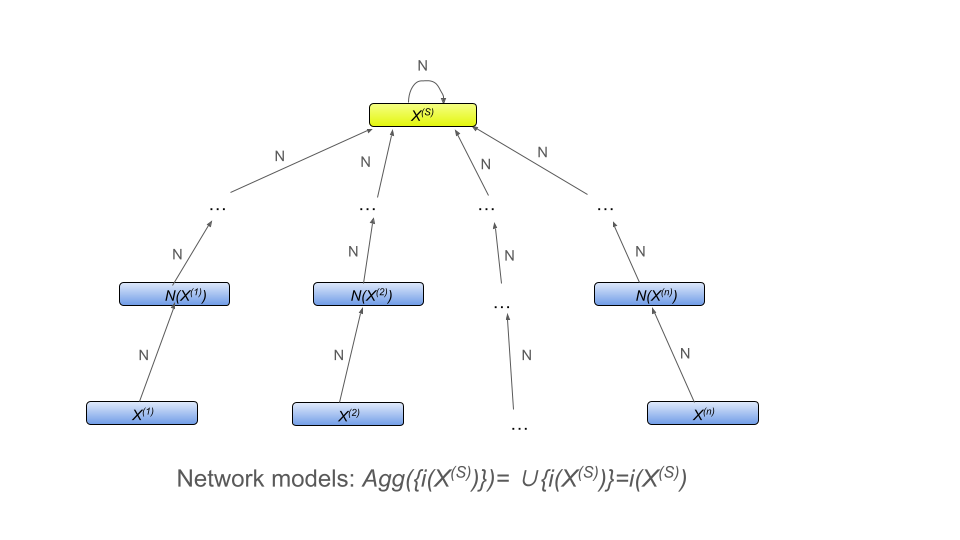}
    \caption{State transition diagram for CORE techniques. The stable state is highlighted in yellow and the corresponding models are given by the stable states.}
    \label{fig:core}
\end{figure}
\noindent Acyclic logic programs are very restrictive as a knowledge representation language. Building on KBANN and the CORE method, CILP is arguably the first neuro-symbolic system for both learning and reasoning \cite{CILP}. Like KBANN, CILP is end-to-end differentiable. Like CORE, CILP is based on recurrent neural networks to work with general or extended logic programs. Instead of introducing multiple hidden layers like KBANN, CILP and CORE use a single-hidden layer neural network in which both the input and output layers represent atoms contained in the logic program. The number of neurons in the hidden layer is equal to the number of clauses in the program. The idea is that the input layer represents an initial (partial) assignment of truth-values to the atoms in the program, while the output layer represents the assignment of truth-values to the atoms that result from applying $T_P$, given the initial assignment. The values of the output layer are then fed back into the corresponding input-layer neurons via recurrent connections, allowing the process to repeat until convergence, that is, allowing the computation of $T_P$ to iterate until a stable state is obtained. We refer to a network of this type as a \textit{cyclic feed-forward network}. As mentioned, encodings using CILP are not restricted to Horn clauses and the atoms in the body of a clause may be negated. It is known that for such (general or extended) logic programs, $T_P$ might not converge. When it does, however, CILP can be used to define a semantic encoding. In such cases, CILP encodings extend the results from the CORE method to differentiable neural networks; see \cite{NeSybook1} for details. 

Fitting CILP into our framework is straightforward with the help of Theorem \ref{lemma:tp}. First, we define our candidate network: the atoms in the body of clauses are mapped to neurons in the input layer, the atoms in the head of clauses are mapped to neurons in the output layer, a hidden neuron is added for each clause. We set $t_c=3$. This means that $N(x)$ will update the values of the neurons three times. The visible units are the input and output neurons (if an atom appears in both the input and output neurons then the corresponding output neuron is made hidden and the input neuron is visible, at the stable state both values will agree) We define our encoding function $i\in I_{NAT}$ by mapping the value of each visible neuron to a truth-assignment of the corresponding atom . The iterative application of $T_P$ requires the definition of an interval for the calculation of threshold value $T$, $0<T<1$, such that for each neuron $k$, the activation value $Act_k$, $-1<Act_k<1$, maps to 1 denoting $true$ if $Act_k>T$, and it maps to $-1$ ($false$) if $Act_k<-T$; otherwise, the truth-value of the atom associated with neuron $k$ is said to be unknown, a third truth value (see \cite{CILP} for details and examples of the translation algorithm from logic programs to neural networks and the computation of $T_P$). Finally, setting $Agg=\cup$ allows us to state the following proposition for general logic programs, that is, programs with cycles and negation by failure.
\begin{proposition}
\label{prop:CILP}
\textit{Given a general logic programming language $\mathcal{S}$ with $P$ a general logic program, if $T_P$ converges to a fixed-point then there exists a cyclic feed-forward network, $N$, that is a semantic encoding of $P$ under $I_{NAT}$ and $Agg=\cup$.}
\end{proposition} 
\begin{proof}
CILP defines the network architecture for a neural network $N$ and proves that $i(N(x))=T_P(i(x))$ in Theorem 8 of \cite{NeSyBook} using the visible-hidden partition, $t_c$, and $i$ as mentioned above. Because every atom in $P$ has a corresponding neuron in $N$, every truth-assignment of the atoms in $P$ is mapped to by a state of the network. Furthermore, by assumption, $T_P$ always converges to a fixed-point. Thus, by Theorem \ref{lemma:tp}, $N$ is a semantic encoding of $P$ under $I_{NAT}$ and $Agg=\cup$.
\end{proof}
The basic construction of CILP has been used to prove similar results for many variants of logic programming, including extended logic programming (allowing negation in the head of clauses), modal, temporal, epistemic and intuitionistic logic programs \cite{NeSyBook}. In all cases, like CILP, neural networks are defined that implement a logical consequence operator with an encoding in $I_{NAT}$. It is a straightforward exercise to duplicate Proposition \ref{prop:CILP} for all of these cases making them all semantic encodings according to our framework. 

Other CORE-like neural encodings exist, e.g. \cite{CORE++,multivalue}. Theorem \ref{lemma:tp} gives a natural way to translate existing results of this type into our framework, with the details in each case being the set of candidate maps and the underlying candidate network (i.e. the partition of neurons into visible-hidden and the computation time $t_c$). Figure \ref{fig:core} shows a general schematic for CORE methods. No matter which initial state, eventually the network converges to a unique stable state which maps to the model of the logic program, given a well-founded logic program, that is, a logic program that admits a single model. Before continuing, we note that not all encodings with logic programs follow the same general formulation used here. In \cite{original2}, an approximate encoding for first-order logic programs is proposed. The programs are acyclic with respect to a given injective level mapping onto feed-forward neural networks, mapping truth assignments to real numbers. Based on \cite{mlpuniv} and using the Banach contraction theorem, a neural network is constructed which implements an approximation of the least-fixed point operator of the logic program. The main theorem of \cite{original2} shows that the distance between the network computation and the least-fixed point operator can be made arbitrarily small. It is straightforward to define an appropriate fidelity measure to show that this is an approximate semantic encoding in our framework, although one with an encoding function that is uncommon, namely the association of interpretations to real numbers. Other more recent approaches and extensions of CILP to first-order logic programs include \cite{CILP++} and \cite{inoue}. The provision of semantic encoding proofs for these systems is left as future work, although the majority of the recent systems rely on grounding every predicate into network embeddings, which should reduce the proofs to a variation of Proposition \ref{prop:CILP}.
\subsection{Penalty Logic}
\label{survey:SCN}
When a class of neural networks is semantically equivalent to a logical system, every neural network in that class can be represented by a knowledge-base of the logical system and vice-versa. Next, we investigate a classic example of semantic equivalence, Penalty Logic \cite{penalty}, before discussing recent work based on graph neural networks and transformers. 

Penalty Logic is a non-monotonic logical system designed to resemble the type of reasoning found in neural networks. In classical propositional logic, if a knowledge-base contains a contradiction then the entire knowledge-base maps to \emph{False}. Penalty Logic eases this restriction by considering sentences in a knowledge-base as evidence instead of facts. To do this, it assigns a confidence value to each sentence in the knowledge-base. It then uses the confidence values to weigh the evidence and come to a conclusion corresponding to the highest confidence out of all possibilities. Formally, the language of Penalty Logic consists of sentences of the form $c : l$, where $c\in \mathbb{N}\cup \{\infty\}$ is a confidence value and $l$ is a sentence in propositional logic. The confidence represents the strength of the evidence for the sentence. It can also be seen as the cost you pay for contradicting $l$. The interpretations of Penalty Logic are defined from the interpretations of propositional logic. Given a knowledge-base $L$ and an interpretation of propositional logic, $M$, we can calculate a penalty, $p_L(M)$, defined as the sum of the confidence values of sentences contradicted by $M$, that is $\sum_{l\in L, M(l)=false} l(c)$ where $l(c)$ is the confidence value of the sentence $l$. The models of a knowledge-base are the interpretations with the minimum penalty. For example, given the sentences $A \vee B$ with confidence $c_1$ and $\neg B$ with confidence $c_2$, the interpretation $(\neg A, \neg B)$ will be assigned penalty $c_1$, interpretation $(\neg A, B)$ will be assigned penalty $c_2$, interpretation $(A, \neg B)$ will be assigned a penalty of $0$, and interpretation $(A, B)$ will be assigned penalty $c_2$. Therefore, $(A, \neg B)$ is the model of the knowledge-base, which is written in Penalty Logic as $\{c_1:A \vee B; c_2:\neg B\}$. 
\begin{figure}
    \centering
    \includegraphics[scale=0.4]{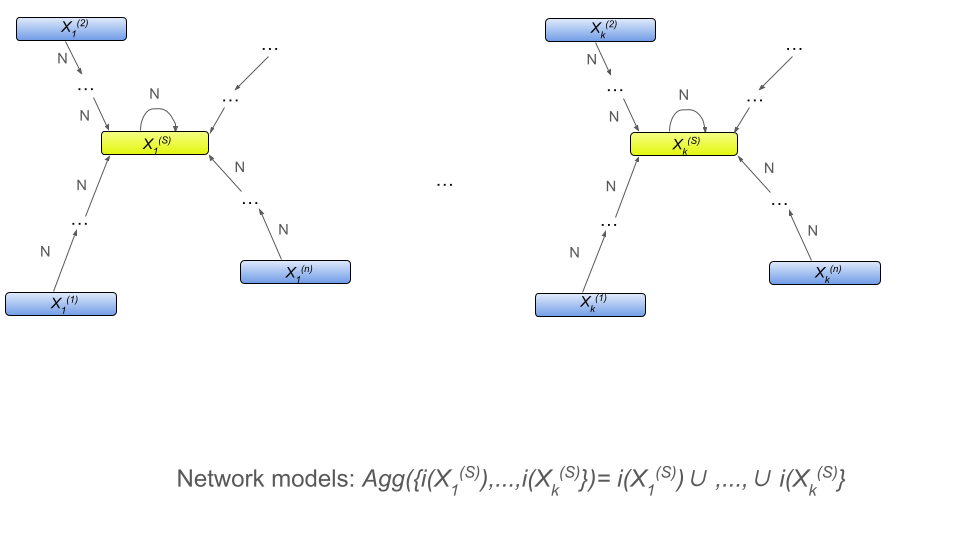}
    \caption{State transition diagram and corresponding models for a Hopfield network encoding Penalty Logic. Stable states are shown in yellow.}
    \label{fig:hopfield}
\end{figure}
Penalty Logic was designed specifically to model the behaviour of a class of recurrent neural networks known as symmetrically-connected network (SCNs), and in particular, Hopfield networks \cite{homer}. SCNs are those in which $w_{i,j}=w_{j,i}$ for all neurons $i,j$. Hopfield networks are binary-valued SCNs in which all transfer functions are the step function and for all neurons $i$, $w_{i,i}=0$. The dynamics of Hopfield networks can be shown to be governed by an \textit{energy function}, i.e. a function $f:X\rightarrow [0,\infty)$ with the property that for all $x\in X$, $f(N(x))\leq f(x)$. It can be shown using this energy function that Hopfield networks are stable and $X_{\textit{inf}}$ consists of the set of states which are local minima of the energy function \cite{hopfield}.\footnote{Note that convergence of the Hopfield network to a stable state depends on the \textit{asynchronous} updating of neurons, that is, updating the values of neurons one at a time. To accommodate this, we can either directly allow for asynchronous updating of neurons in our framework, or define a probability distribution which selects, with a certain probability, which neuron to update at each time step. In the later case, the encoding in Proposition \ref{prop:Hopp} will become a probabilistic semantic encoding.} In \cite{penalty}, equivalence is shown between Hopfield networks and knowledge-bases in Penalty Logic by showing that, with an encoding function $i\in I_{NAT}$, the energy function of every Hopfield network corresponds to a penalty function $p_L$ of a knowledge-base $L$, and vice-versa. We use this to prove the following proposition.
\begin{proposition}
\label{prop:Hopp}
If a Hopfield network, $N$, always converges to a global minimum of its energy function then it is a semantic encoding of a knowledge-base of Penalty Logic $L$, under $I_{NAT}$ and $Agg=\cup$.
\end{proposition}
\begin{proof}
 In \cite{penalty}, a propositional atom is associated with each neuron in $N$. Each state of the neural network is associated with an assignment of truth-values to these atoms, $i_0$, in the usual way with $i_0(x)$ mapping each atom to true iff the corresponding neuron has a value of $1$. A knowledge-base, $L$, is constructed with the set of atoms corresponding to the neurons of $N$ and Theorem 4.10 in \cite{penalty} shows that $E(x)<E(x')$ if and only if $p_L(i_0(x))<p_L(i_0(x'))$. In particular, the global minima of $E$ correspond to models of $L$. We define our encoding function $i\in I_{NAT}$ by mapping each state, $x$, to the set of models in which the truth-values of the atoms contained in $L$ agree with those given by $i_0(x)$. It is easily shown that for all $M\in\mathcal{M}$, there exists $x\in X$ with $M\in i(x)$ by taking a state, $x$, in which $i_0(x)$ agrees with $M$ for all atoms in $L$. This state must also be unique as every other state must assign at least one atom in $L$ to a different truth-value. 
 
 If $M,M'\in i(x)$ then, because they must agree on all atoms in $L$, $p_L(M)=p_L(M')=p_L(i_0(x))$. By assumption, if $x\in X_{\textit{inf}}$ then $E(x)\leq E(x')$ for all $x'$ and thus if $M\in i(x)$, we have that $p_L(M)=p_L(i_0(x))\leq p_L(M')$ for all $M'\in \mathcal{M}$. Thus, $M$ is a model of $L$. Conversely, let $M$ be a model of $L$ and $x\in X$ be the state with $M\in i(x)$. For all $x' \in X$ with $M'\in i(x')$, because $M$ is a model of $L$, $p_L(M)\leq p_L(M')$, which implies that $E(x)\leq E(x')$. Thus, $x\in X_{\textit{inf}}$ meaning $\bigcup\limits_{x\in X_{inf}} i(x) =\{M| M\in i(x), x\in X_{inf}\}=\mathcal{M}_L$. Hence $\mathcal{M}_N=\mathcal{M}_L$ so $N$ is a semantic encoding of $L$ under $I_{NAT}$ and $Agg=\cup$. 
\end{proof}
Theorem 4.10 in \cite{penalty} also proves that for every knowledge-base $L$, there exists a Hopfield network (possibly with hidden units) with $\bar{E}(x)\leq \bar{E}(x')$ iff $p_L(i_0(x))\leq p_L(i_0(x'))$, where $\bar{E}(x)$ is defined as the minimum energy of all states in which the state of the visible units is identical to $x$. What keeps us from using this to prove semantic equivalence between Hopfield networks and Penalty Logic is the existence of local minima in the general case. If $x\in X_{\textit{inf}}$ in a Hopfield network, it is not always true that $x$ is a global minimum of $E$ and thus it is not always true that $i(x)\subset \mathcal{M}_L$. The existence of local minima is seen as undesirable and there are methods for addressing this such as simulated annealing which can guarantee convergence to global minima. In the case that the Hopfield network corresponding to $L$ does not have local minima then the previous proof can be used to show that the network is a semantic encoding of $L$ under $I_{NAT}$ and $Agg=\cup$. Figure \ref{fig:hopfield} shows this relationship visually. Here we can see that each starting state eventually converges to a stable state each of which maps to a set of models of a penalty logic knowledge-base. Aggregating these sets with a union results in the entire set of models for the knowledge-base.

\subsection{Logical Classifiers}

Recent work has also been interested in finding equivalence between modern feed-forward networks and functions expressed using logic. Here the neural network is seen as a function mapping an input vector to an output label, and the desire is to find a logical expression that correspond to this function. We show in this section that such network representations as logical expressions can also be viewed as examples of semantic encoding. We start by looking at a case involving Graph Neural Networks (GNNs).

The computational power of GNNs has been the subject of much inquiry. In \cite{GNN:semantic_equiv}, it is shown that Boolean classifiers, i.e. functions classifying nodes in graphs as true or false, expressible as formulas in the fragment of first-order logic with two-variables and with counting quantifiers can be learned by GNNs. A counting quantifier, $\exists^{\geq N}x(P(x))$, allows one to express that a formula, $P(x)$, holds true for at least $N$ instances of the variable, $x$. The fragment of first-order logic with two variables and counting quantifiers is known as the FOC$_2$ fragment.


First let us define the GNNs investigated in \cite{GNN:semantic_equiv}. GNNs are a generalization of classical neural networks that operate over graph embeddings in $\mathbb{R}^n$. A graph, $\mathcal{G}$, in this case, is a finite set of nodes $V$ with a symmetric relation, $E\subseteq V\times V$. A labeled graph is a graph $\mathcal{G}$ in which each node, $v$, is associated with a vector $x_v \in \mathbb{R}^k$ for some $k \geq 1$. An Aggregate-Combine GNN (AC-GNN) transforms a labeled input graph into a sequence of labeled latent graphs according to the following formula:
\begin{equation}
    x_v^{(i)} = COM^{(i)}\left( x_v^{(i-1)}, AGG^{(i)}(\{x_u^{(i-1)}| (u,v) \in E\})\right)
\end{equation}
where $i$ is the $i^{th}$ layer, $AGG^{(i)}$ is the $i^{th}$ aggregation function which takes as input a set of vectors of neighbouring nodes and returns an output vector, and $COM^{(i)}$ is the $i^{th}$ combination function which combines the aggregated features in the previous layer of the neighbouring nodes with the node features from the previous layer of the target node to output a new feature vector for the target node. The final layer assigns a label to each node. Given an input graph, a state of a GNN is an assignment of vector values to each node in each layer of the GNN. Notice that the dimension of the vector for each layer is fixed, so if an input graph has $n$ nodes and the GNN maps the $m^{th}$ layer to vectors in $\mathbb{R}^k$ then the state of the nodes in the $m^{th}$ layer is given by a vector of length $\mathbb{R}^{nk}$. Given a logical system, $\mathcal{S}=(\mathcal{L},\mathcal{M})$, for a given input graph, an encoding function $i_{V,E} \in I$ is a function from the state space of the GNN with input graph $(V,E)$ to $2^\mathcal{M}$. For each input graph, $(V,E)$, let $X_{(V,E)}$ be the state space of the GNN with input graph $V,E$. Define $X_{inf,VE}$ to be $\{x \in X_{(V,E)} | \exists x_0 \in X_{(V,E)}, \lim\limits_{t\rightarrow \infty} GNN_{V,E}^{(t)}(x_0) = x\}$, where $GNN_{V,E}^{(t)}(x_0)$ is the result of updating the state according to Eq.(2) $t$ times. Finally, given an aggregation function $Agg:2^{2^\mathcal{M}}\rightarrow 2^\mathcal{M}$, let $\mathcal{M}_{GNN}=Agg(\{i_{V,E}(x)| x \in X_{inf,(V,E)}, (V,E) \in \mathcal{G}\}$. Then, the definitions of a neural model and semantic encoding for a GNN are identical to the ones found in Definition $\ref{semanticneuralmodel}$.
In \cite{GNN:semantic_equiv}, it is proved that AC-GNNs can learn functions on graphs expressed using a fragment of first-order logic known as graded modal logic, where all sub-formulas are guarded by the \emph{edge} predicate $E(x,y)$. Instead of writing $\exists x(P(x))$ to denote that some node $x$ in the graph has property $P$, such as e.g. the colour of the node, the guarded version $\exists x(E(y,x) \wedge P(x))$ has to be used, denoting that $P$ holds for a node within a neighbourhood defined by the edges that exist between node $y$ and other nodes. Also, an extension of AC-GNNs called ACR-GNNs, standing for Aggregate-Combine-Readout, in which in each layer a feature vector for the entire graph is calculated and combined with local aggregations, is shown to be capable of representing FOC$_2$ \cite{GNN:semantic_equiv}, in the sense that for a given input graph and initial state, the output label of each node calculated by the GNN can be expressed by a formula of the form $\alpha(x) \leftrightarrow \phi(x)$, where $\alpha$ is a unary predicate representing the label of a node and $\phi$ is a sentence in FOC$_2$. The domain of these predicates are sets of nodes in a graph. The truth values of the predicates in $\phi$ are determined by the graph structure and additional \textit{input labellings}. For example, $E(x,y)$ is true if and only if there is an edge between x and y, whereas an additional predicate $B(x)$ representing e.g. that the colour of a node is \textit{blue} is true only if it is given in the input labelling. This means that given a graph with a set of input labellings for the node, the final layer of the GNN will assign an output label $
\alpha$ to the node $x$ if and only if the input graph satisfies $\phi(x)$.

The correspondence between GNNs and first-order logic fragments proved in \cite{GNN:semantic_equiv} can be shown to be semantic encodings as outlined next.
\begin{proposition}
Given a logical system whose interpretations consist of graphs and whose predicates consist of node labellings plus the edge relation, for each sentence of the form $L=\forall x (\alpha(x) \leftrightarrow \phi(x))$, if $\phi(x)$ is expressed in graded modal logic then there exists an AC-GNN that is equivalent to $L$. If $\phi(x)$ is expressed in FOC$_2$ then there exists an ACR-GNN that is equivalent to $L$.
\end{proposition}\label{gnn}
\begin{proof}
For a given $GNN$, let $i_{V,E}(x)$ denote the set of interpretations in which $E(x,y)$ is true iff $x$ is connected to $y$,$\alpha(x)$ is true if and only if the output layer for node $x$ has label $\alpha$, and for any remaining predicates, $P(x)$ is true if the input layer of node $x$ has label $P$. Let $Agg=\cup$. Then, for a given labeled input graph, the stationary state is the one in which the GNN has propagated the input labeling through each layer up to the output. If a GNN can be equivalently described by a Boolean classifier then in the stationary state, for each node $x$, $i_{V,E}$ maps $\alpha(x)$ to true if and only if it maps $\phi(x)$ to true.  Hence, the set of stationary points for a GNN is the set of points that are models of $\alpha(x) \leftrightarrow \phi(x)$ and vice-versa, making the GNN a semantic encoding of the knowledge base.
\end{proof}

This type of equivalence involves mapping a Boolean classifier defined by a logical formula $\phi$ onto a neural network by mapping assignments of truth values and predicates to patterns of input and output neurons. In these equivalences, a neural network maps a pattern of input neurons to a truth assignment of ground predicates, and output neurons to an output label. The network is said to be equivalent to the Boolean classifier if the output label is $1$ if and only if the truth assignment of the input pattern satisfies $\phi$. This type of correspondence can in fact be shown to always represent a semantic encoding. Next, we formalize this for the case without free variables; a proof of the general case should remain valid but would contain a lot more bookkeeping.

First, we formally define a logical classifier over a language, $\mathcal{S}=(\mathcal{L},\mathcal{M})$, as a function $f:\mathcal{M}\rightarrow \{0,1\}$ with $f(M)=1$ if and only if $M$ is a model of some sentence $\phi$. We say that a neural network encodes this classifier if the following holds
\begin{definition}
    Let $N$ be a feed-forward neural network with $X_{in}$ the state space of the input neurons, $X_{out}$ the state space of the output neurons, and let $g_{in}:X_{in}\rightarrow \mathcal{M}$, $g_{out}:X_{out}\rightarrow \{0,1\}$. If $f$ is a logical classifier over $\mathcal{S}=(\mathcal{L},\mathcal{M})$, we say that $N$ implements $f$ if $g_{in}$ is surjective and $g_{out}(N(x_{in}))=1$ iff $f(g_{in}(x_{in}))=1$, where $N(x_{in})\in X_{out}$ is the state of the output neurons after propagating the input pattern $x_{in}$ through all intermediate hidden layers.
\end{definition}
This definition is intuitively simple: a feed-forward neural network acts as a function from its input state to its output state. If the input state represents interpretations of a logical system and the output state represents the outcome of a classifier then the neural network implements a logical classifier if the function computed by the network is the function $f$ that defines the classifier. For the case in which $\mathcal{S}$ is a first-order logic system and the classifier has free-variables, the situation is more or less identical except that the output of the network must represent a mapping from the grounding of the formula containing free-variables to truth values. This is the case we just examined with AC(R)-GNNs, but to keep the following proof straightforward, we stick to the no-free variable case.

The set of logical classifiers over a language $\mathcal{S}=(\mathcal{L},\mathcal{M})$ defines a new logical system $\mathcal{S}_C=(\mathcal{L}_C,\mathcal{M}_C)$ in which $\mathcal{L}_C$ consists of sets containing a single sentence of the form $\alpha \leftrightarrow \phi$ where $\phi \in \mathcal{L}$ and $\alpha$ is a new nullary predicate. Interpretations $\mathcal{M}_C$ consist of an interpretation $M\in \mathcal{M}$ that additionally assigns a truth value to $\alpha$. For simplicity, we denote $M_C$ by a pair $(M,\hat{\alpha})$ where $M\in\mathcal{M}$ and $\hat{\alpha}\in \{0,1\}$ represents the truth assignment to $\alpha$ ($1$ for $\alpha=True$, $0$ for $\alpha=False$). $M_C\in \mathcal{M}_C$ is a model of $\alpha \leftrightarrow \phi$ if either the corresponding $M\in \mathcal{M}$ is not a model of $\phi$ and $M_C$ assigns $\alpha$ to false, or $M\in \mathcal{M}$ is a model of $\phi$ and $M_C$ assigns $\alpha$ to true. Clearly, there is a one-to-one correspondence between logical classifiers and $\mathcal{L}_C$ in which each classifier, $f$, is associated with the knowledge-base $L_f=\{\alpha\leftrightarrow \phi_f\}$ and the models of $L_f$ are all pairs $(M,\hat{\alpha})$ where $\hat{\alpha}=1$ if $f(M)=1$ and $\hat{\alpha}=0$ if $f(M)=0$. This leads to the following theorem.

\begin{theorem}\label{lc}
    $N$ implements a logical classifier $f$ if and only if $N$ is a semantic encoding of the corresponding knowledge-base $L_f=\{\alpha\leftrightarrow \phi_f\}$ under $i(x)=(g_{in}(x_{in}),g_{out}(x_{out}))$ and $Agg=\cup$.
\end{theorem}
\begin{proof}
    By assumption $N$ is feed-forward and all neurons that are not in the input or output layers are hidden. This means that for each input pattern $x_{in}\in X_{in}$, $N$ has a unique fixed point at $(x_{in},N(x_{in}))$. This fixed point corresponds to the interpretation $i(x)=i((x_{in},N(x_{in})))=(g_{in}(x_{in}),g_{out}(N(x_{in})))=(M,g_{out}(N(x_{in})))$. If $N$ implements a logical classifier then $g_{out}(N(x_{in}))=1$ iff $f(M)=1$. Thus, $(M,g_{out}(N(x_{in})))$ is a model of $L_f$. Conversely, if $(M,\hat{\alpha})$ is a model of $L_f$ then $f(M)=\hat{\alpha}$ and since $N$ implements $f$, $g_{in}$ is surjective. Thus, $M=g_{in}$ for some $x_{in}$ and because $N$ implements $f$, $g_{out}(N(x_{in}))=\hat{\alpha}$. Hence, the fixed-points of $N$ correspond exactly to the models of $L_f$ meaning that $N$ is a semantic encoding of $L_f$ under $i$ and $\cup$.\\
    Now, suppose that $N$ is a semantic encoding of $L_f$ under $i$ and $\cup$. For each fixed point, $(x_{in},N(x_{in}))$, $i(x_{in},N(x_{in}))=(g_{in}(x_{in}),g_{out}(N(x_{in})))$ is a model of $L_f$ and all models of $L_f$ are of this form. Because all models of $L_f$ are of this form, $g_{in}$ must be surjective. Furthermore, because this must be a model of $L_f$, $g_{out}(N(x_{in}))=1$ iff $f(g_{in}(x_{in}))=1$ and thus $N$ implements $f$.
\end{proof}

In the case that a logical classifier contains free-variables, an analogous proof exists. The main difference is that the output layer of the feed-forward network should now represent a mapping from variable assignments to truth values rather than a single truth value. As discussed, we provided an example of this case earlier in which the value of each node in the output graph represents the truth assignment for a predicate that is grounded by that node. Theorem \ref{lc} allows us to substantially simplify the proof of correspondence with logical classifiers to semantic equivalences, as we illustrate with the next proposition.
\begin{proposition}\label{transformer}
    A transformer is a semantic encoding of a knowledge-base in first-order logic extended with majority quantifiers.
\end{proposition}
\begin{proof}
    In \cite{transformer_equiv}, it is shown that transformers implement logical classifiers that can be described using first-order logic extended with the majority quantifier, \textbf{M$_i$}. A formula \textbf{M}$_i\phi_i$ is true if $\phi_i$ is true for more than half of the possible values of variable $i$. Interpretations consist of strings with a finite number of characters (or tokens).\footnote{In a language consisting of tokens $a$ and $b$, for example, \textbf{M}$_ib(i)$ denotes the strings with more $b$'s than $a$'s. We say that token $i$ is $b$. In the case of strings with three tokens, the models of \textbf{M}$_ib(i)$ are $abb$, $bab$, $bba$ and $bbb$.} From Theorem \ref{lc} it follows directly that transformers are semantic encodings of such logical classifiers.
\end{proof}

Propositions \ref{gnn} and \ref{transformer} show that results obtained independently about the correspondence between a given logical formalism and a class of neural networks can be unified, represented uniformly and subsumed by the framework of semantic encoding. 

Many modern approaches do not aim for an exact representation (e.g. showing that a transformer can be equivalently represented by sentences in first-order majority logic), but instead use the loss function to train a network to approximate a knowledge base. we now turn our attention to these cases.
 
\label{sec:semantic_reg}
\subsection{Semantic Regularization}

As we have seen, traditional approaches to neuro-symbolic computation generally attempt to encode a knowledge-base into a neural network exactly. However, as discussed in Section $\ref{sec:approx_encode}$, many recent methods have sought to encode a knowledge-base into the loss function, which is to be minimized via learning. In the language of our framework, these loss functions are fidelity measures. In other words, they measure how far a neural network is from being a semantic encoding of a given knowledge-base. These techniques may broadly be called \textit{semantic regularization} techniques, in that the knowledge-base acts as a regularizer term in the loss function. In this section, we review some prominent semantic regularization techniques.

We begin with a family of encodings in $I_{DAT}$ that maximize $Fid_{fuzzy}$. In these methods, a neural network is used to implement for each predicate $P$, a function $F_P:O^k\rightarrow [0,1]$, where $O$ is a set of objects and $k$ is the arity of $P$, and for each function symbol, $f$, a function $F_f:O^k \rightarrow O$, where $k$ is the arity of $f$. The network then defines a fuzzy interpretation of the language in which the domain of discourse is $O$, a predicate $P(o_1,...,o_k)$ is given the truth value assigned to it by $F_P(o_1,...,o_k)$, and the truth value of a sentence, $\phi$, is determined recursively using chosen fuzzy connectives, quantifiers and the neural interpretations of the function symbols and constants. Given a knowledge-base, $L$, the network is trained to maximise satisfiability of $L$. We show that this process is equivalent to maximizing $FID_{fuzzy}((N,i,\cap),L)$ where $i\in I_{DAT}$. A general account of semantic regularization methods can be found in \cite{diff_fuzz}. In what follows, we will use Logic Tensor Networks (LTNs) \cite{LTN} as our prototypical example.

LTN modifies first-order logic so as to align the operations used by an interpretation more closely to those used by neural networks. The result is \textit{Real Logic}, a many-valued logic whose sentences are sentences in a first-order logic with semantics defined by a domain $O\subset \mathbb{R}^n$, fuzzy operators, and \textit{groundings} of first-order sentences. A grounding interprets the constants, $c$, functions, $f$, and relations, $R$, of a first-order language to define an embedding in the following way:
\begin{itemize}
    \item $G(c)\in O$ for every constant symbol $c$;
    \item $G(f) \in O^{\alpha_f}\rightarrow O$ for every function symbol with airity $\alpha_f$;
    \item $G(R) \in O^{\alpha_R}\rightarrow [0,1]$ for every predicate $R$ with airity $\alpha_R$.
\end{itemize}
We can combine these to define a mapping, $G$, of terms into tensors, which can be used along with the chosen fuzzy operators to define a mapping from sentences to truth values in the interval $[0,1]$.
\begin{itemize}
    \item $G(f(t_1,...,t_m))=G(f)(G(t_1),...,G(t_m))$;
    \item $G(R(t_1,...,t_m))=G(R)(G(t_1),...,G(t_m))$;
    \item $G(\neg l)=1-G(l)$;
    \item $G(l_1\vee ... \vee l_k)=\mu(G(l_1),...,G(l_k))$, where $\mu$ is a fuzzy t-conorm (e.g. $\max$);
    \item $G(\forall x l(x))= A_{(x_0\in O)} l(x_0)$, where $A$ is a fuzzy universal quantifier.
\end{itemize}
Notice that when $O$ is finite, we can use functions such as $\sum$ or $\min$ as $A$; otherwise, in practice, $A$ will be an approximation of a fuzzy universal quantifier.

Groundings define the interpretations of Real Logic. 
The groundings and interpretations of Real Logic are in a one-to-one correspondence and as such we will not distinguish between them from this point on. 
Intuitively, a constant is a feature vector grounding an object onto its properties, such as colour represented by its RGB values. It is helpful to define a notion of a partial grounding, $\hat{G}$, which is a grounding that is only defined for some terms in the signature. Given a knowledge-base, if a partial grounding is defined on all terms and predicates that appear in the knowledge-base then we can determine whether or not the partial grounding satisfies the knowledge-base. A partial grounding can be seen as an equivalence class of groundings that agree on the terms and predicates on which the partial grounding is defined. A Logic Tensor Network is simply an implementation of a partial grounding with a neural network. The goal is, given a knowledge-base, $L$, to have the neural network learn a partial grounding that satisfies $L$. One can see that if an LTN is trained successfully then it defines a neural model on $L$. The neural network represents a partial grounding that satisfies $L$, which in turn represents a set of groundings that satisfy $L$, which in turn represents a set of interpretations that satisfy $L$. Next, we show how an LTN successfully trained on a knowledge-base $L$ represents in our framework a neural model of the training set under $I_{DAT}$ and $Agg=\cap$.
\begin{figure}
\centering
\subfloat[LTN architecture]{
\begin{tikzpicture}[scale=0.7]


 \node[shape=circle,draw=black, anchor=center,minimum size=0.01cm, label=below:$\bar{x}_{1,1}$] (a1) at (-2,1){} ;
 \node[shape=circle,draw=none, anchor=center,minimum size=0.8cm] (a2) at (-1,1){...} ;
 
  \node[shape=circle,draw=black, anchor=center,minimum size=0.01cm, label=below:$\bar{x}_{1,n}$] (a3) at (0,1){};
  
 \node[shape=circle,draw=none, anchor=center,minimum size=0.8cm] (mid1) at (2,1){...} ; 
  
  \node[shape=circle,draw=black, anchor=center,minimum size=0.01cm,label=below:$\bar{x}_{m,1}$] (b1) at (4,1){};
 
 \node[shape=circle,draw=none, anchor=center,minimum size=0.8cm] (b2) at (5,1){...} ;
 
 \node[shape=circle,draw=black, anchor=center,minimum size=0.08cm,label=below:$\bar{x}_{m,n}$] (b3) at (6,1){};
 
 \node[trapezium, trapezium angle=60, minimum width=50mm, draw, thick] (t1) at (2,2.5) {Hidden Neurons};
 
  \node[shape=circle,draw=black, anchor=center,minimum size=0.08cm,label=left:$x_{(R\theta_1,...,\theta_k)_1}$] (c1) at (0,4){} ;
  
 \node[shape=circle,draw=none, anchor=center,minimum size=0.8cm] (mid2) at (2,4){...} ; 
  
  \node[shape=circle,draw=black, anchor=center,minimum size=0.08cm,label=right:$x_{(R\theta_1,...,\theta_k)_l}$] (c2) at (4,4){};

\draw[->] (a1)--(t1);
\draw[->] (a2)--(t1);
\draw[->] (a3)--(t1);
\draw[->] (mid1)--(t1);
\draw[->] (b1)--(t1);
\draw[->] (b2)--(t1);
\draw[->] (b3)--(t1);
\draw[->] (t1)--(c1);
\draw[->] (t1)--(c2);
\draw[->] (t1)--(mid2);

\end{tikzpicture}
}
\hfill
\subfloat[State transition diagram and corresponding models]{\includegraphics[scale=0.4]{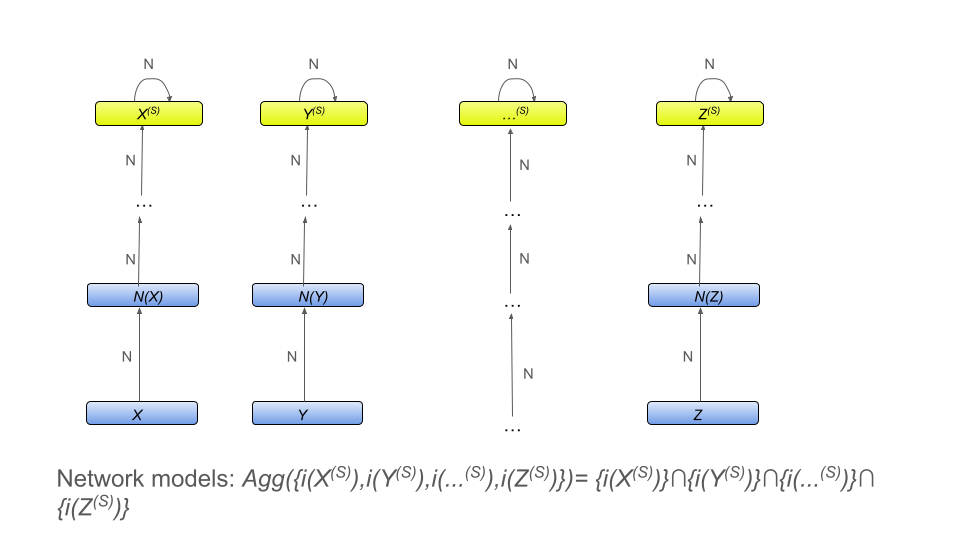}}

\caption{The general structure of an LTN (a), where the neural network can be any architecture, a GNN, Transformer, feedforward, recurrent network, etc. The state transition diagram for the LTN (b). Note that the state-space is uncountable. We show the input states as a sequence merely for visualization purposes.}
\label{nnfig:6}
\end{figure}
\begin{proposition}
    Minimizing the loss function of an LTN, $N$, on a knowledge-base, $L$, is equivalent to maximizing $Fid_{fuzzy}((N,i,\cap),L)$ with $i\in I_{DAT}$.
\end{proposition}
\begin{proof}
See Appendix C.
\end{proof}

LTNs are a prominent representative of a family of semantic regularization techniques. They differ in the choice of fuzzy connectives and the restrictions imposed on the type of sentences in the language \cite{sem_reg,diff_fuzz}. In our framework, all of these techniques can be summed up as encodings under $I_{DAT}$ which are trained to maximize $Fid_{fuzzy}$. This opens the door for a general formalization of these techniques without committing to a single method. Our framework also extends to semantic regularization methods that do not use fuzzy logic but use probability theory, as we show next.

In \cite{sem_loss}, a generic semantic loss function for deep neural networks is proposed whose output labels represent propositional variables. Each labeling predicted by the neural network represents an assignment of truth values to the propositional variables. Each neuron in the output layer is assumed to represent the probability that the variable corresponding to that neuron is true. Then, the semantic loss is defined by $Loss(L,p)=-\log \sum\limits_{M:M(L)=True} \prod\limits_{j: M(X_j)=True} p_j \prod\limits_{j: M(X_j)=False} (1-p_j)$, where $L$ is a propositional knowledge-base, $p$ is the vector with the probabilities that each variable represented by the output layer is assigned to true, and $M$ is an assignment of the $n$ propositional variables represented by the output neurons. We show that minimizing this loss is equivalent to maximizing $Fid_{prob}$. 

\begin{proposition}
    Minimizing the semantic loss of a feed-forward network, $N$, to a propositional knowledge-base $L$ is equivalent to maximizing $Fid_{prob}((N,i,\cup),L)$ with $i\in I_{NAT}$.
\end{proposition}
\begin{proof}
See Appendix D.
\end{proof}

By formalizing semantic regularization techniques in the language of semantic encodings, we are able to quickly identify the similarities and differences between such techniques. While we looked at approximate encodings in $I_{NAT}$ with $Fid_{prob}$, other methods combine the distributed encoding of LTN with probabilistic optimization, e.g. \cite{relational_neural_machines}, creating an approach that seeks to maximizes the \textit{expected satisfiability} of a knowledge-base. This is an encoding in $I_{DAT}$ with $Fid_{prob}$. As with all of the methods discussed in this section, this approach weighs the information coming from the observations in the data with a measure of the distance of a neural network from being a model of a knowledge-base. Finally, one could expect to design a neuro-symbolic technique using $I_{NAT}$ and $Fid_{fuzzy}$, or even create a new fidelity measure based on the existing ones and a new set of candidate maps. By systematizing semantic regularization techniques using the language of semantic encodings, we are able to show that, despite offering very different formalizations, they are approximate versions of the classic neuro-symbolic encoding techniques.

\subsection{Other Neuro-symbolic Encodings}

While the techniques that we have discussed represent a significant portion of the literature on neuro-symbolic encoding, there are many techniques that we have not included. Recent interest in neuro-symbolic computing has led to a proliferation of techniques making it impossible to cover them all here. Crucially, not all techniques can be described as semantic encodings. In particular, many methods map states of a neural network directly onto sentences of a logical system rather than interpretations and models (see \cite{tprreason} and \cite{end-to-end} for examples). To differentiate, we call these \textit{syntactic approaches}, outside of the semantic framework proposed here. In the closely-related area of neurosymbolic programming \cite{NP}, for example, it is worth also distinguishing between syntactic approaches and the semantics of program synthesis. Notably, however, \cite{inhibitionnets} define an encoding in which each sentence in the language is mapped to a state of the network. A method is described that associates to each of these states a mapping from neurons to interpretations. It should be possible to extend this idea to encoding functions, suggesting a relationship between syntax based encodings of this kind and the type of semantic encoding described by our framework. 

The semantic framework, therefore, covers those methods in which states of a neural network are mapped onto interpretations of a logical system. We have seen that the proposed definitions capture a large number of neuro-symbolic techniques. Although the formalization in the framework of other techniques not investigated here is left as future work, in some cases the relationship should be obvious. For example, Proposition 4.3 in \cite{MLN} shows that, as the weights of an MLN go to infinity, the probability that a state of the MLN is a model of the desired knowledge-base converges to $1$. This is clearly an approximate encoding under $Fid_{prob}$ according to our definitions. Another recent approach that has become common is to use Inductive Logic Programming with a differentiable structure \cite{LNN,DILP}. In these approaches, a vector space is used to represent truth assignments to ground atoms and the weights between atoms are learned from examples. The resulting structure then implements a logic program as analyzed in Section \ref{survey:LP}. Other examples modify a network to satisfy hard constraints \cite{hard_contraint}. These methods can be seen as a combination of semantic encoding and semantic regularization.

\begin{table*}[htbp!]
\footnotesize
\caption{\label{table:semantic_encoding_comparison}
 Summary of the semantic encoding classes investigated in this paper with their main characteristics: candidate mapping ($i$), aggregation approach ($Agg$) and neuro-symbolic encoding. 
}

\begin{center}
\centering
\begin{tabular}{| c | c | c | c|}
\hline 
 Encoding Method &  $i$ & $Agg$ & Encoding Technique \\ 
 \hline 
 \hline 
 CORE Methods & $I_{NAT}$ & Union & Direct Encoding   \\  
 \hline 
 Penalty Logic & $I_{NAT}$ & Union & Equivalence   \\
 \hline 
 Graph Neural Networks & Unique & Union & Direct Encoding  \\
 \hline 
 Fuzzy Differentiable Operators & $I_{DAT}$ & Intersection & Approximate ($FID_{fuzzy}$)  \\
 \hline
 Semantic Loss & $I_{NAT}$ & Union & Approximate ($FID_{prob}$)  \\
\hline
\end{tabular}
\end{center}

\end{table*}

Another line of work suggests possible extensions of the semantic framework; for example, \cite{HebbianLogic} use neural networks to model a modal logic based on Hebbian learning. Similarly, \cite{Trann:2016} uses Contrastive Divergence, although the approach in \cite{Trann:2016} is squarely a semantic approach. Taking inspiration from this work and possible-worlds semantics, our framework could be extended to sets of neural networks whereby each network represents a possible world and the relationships between the networks are either induced by learning or pre-defined, as done in \cite{NeSyBook}. Table 1 summarises the methods that have been studied in this paper. The encoding function class refers to the set of candidate maps that the encoding function belongs to. It is labeled as \textit{unique} if the encoding function does not belong to a set of candidate maps that we have pre-defined. The aggregation function is in all cases either \textit{union} or \textit{intersection}. The encoding technique is how the network is modified to become a semantic encoding: either the weights are directly calculated to make the network a semantic encoding (direct encoding), the network is trained to approximate a semantic encoding (approximate encoding), or every network in the class is represented by a knowledge-base and every knowledge-base is represented by a network (equivalence). 

We can see from the table that the existing paradigms of semantic mappings (ie directly encoding knowledge bases into the weights of a network) and semantic regularization (using the loss function to steer the network towards models of the background knowledge) fit comfortably into our framework. In the former case, we begin with a network that is a model of the knowledge base and train it with additional data, whereas in the latter, we begin by training it with data while regularizing the loss. The result of this is that semantic mappings start out as neural models pre-training and will diverge as inconsistent data is presented, while in semantic regularization the network starts out far from being a model and gets closer and closer to a neural model as the network is trained. In this sense semantic mappings prioritize the semantic relevance of background knowledge and operate under the assumption that potential contradictions of this knowledge in the training data are minor and that training will not result in the network deviating from the set of neural models too much. Regularization techniques take the opposite approach, they emphasize the flexibility of learning from data and assume that much of the relevant information required to learn a concept is not contained in, and possibly contradicts, the background knowledge. The choice of approach likely depends on the application area and the confidence in the relevance of the background knowledge being applied. Finally, equivalences are a matter of representation, they show that a class of neural networks can be precisely described by a logical system and vice-versa meaning that the tools available for the analysis of each of these can be applied to each other.

Having the language of semantic encoding allows us compare these techniques directly and is expected to help organize the discussion around the common components and properties of this large class of neuro-symbolic techniques. 

\section{Conclusion and Future Work}
 For a long time, it has been recognized by some that rule-based symbolic systems and artificial neural networks have complementary strengths. This has led to the development of a field of AI looking to address its main challenges  through \textit{neural-symbolic integration}. The object of neuro-symbolic AI is to find translation algorithms to and from symbols and networks, showing correspondence and equivalence results, and to develop a learning and reasoning system based on deep learning with symbolic interpretation. The promise of neuro-symbolic AI is to produce, as a result of the combination of data-driven learning and knowledge, more robust (adaptable to new tasks), efficient (learning from fewer examples), and  transparent AI systems (explainable through symbolic computation).
 
 What has been lacking in neuro-symbolic AI is a theory that can be leveraged for the design of new AI systems. In this paper, we have made some first steps towards such a theory by formalizing a framework for an important class of neuro-symbolic techniques that we call \textit{semantic encodings}. This is far from the final word on a theory of neuro-symbolic AI and should be looked at instead as a starting point for work in this area. We hope that this paper can serve both as a formal introduction to neuro-symbolic AI for those who are new to the field and as a guide to help organize with a common context the many different approaches to neuro-symbolic computation. We conclude by outlining various directions for future work which are derived from the contributions of this paper.
 
 The most obvious direction for future work is to continue to expand on the definitions provided here so that they can be used to describe more neuro-symbolic approaches. Within the context of semantic encoding, task-specific definitions of fidelity and descriptions for other useful sets of candidate mappings are desirable. We have also not discussed a number of neuro-symbolic techniques which are not semantic in nature, that we called \textit{syntactic} techniques, e.g. \cite{deepproblog}, and the definition of a framework capable of capturing these and other hybrid systems should also be a priority of future work. Finally, developing a theory stemming from Section 3.4 could provide insight into the effectiveness of neural encoding in practice in both the exact and approximate cases.

 The most important thing going forward for neuro-symbolic AI as a whole, we feel, is to use theoretical properties of semantic encodings to inform the development of new encoding techniques. Rather than develop ad-hoc methods that iterate on previous techniques, a systematic analysis of contemporary techniques such as semantic regularization should take place. This should clarify the role of components such as the encoding function, the fidelity measure, and the background knowledge in producing methods that can deliver on the promises of neuro-symbolic integration. Using the framework to generalize results should as those found in \cite{marconato2023neurosymbolic} should give us insight into the limitations and potential of current techniques and allow us to approach the development of new semantic encodings in an informed way. This will be aided by further development of a neuro-symbolic theory, especially one that focuses on the relationship between encodings and learning along the direction suggested in section 3.4. Finding a way to characterize properties of the encoding function that are required for it to be relevant to learning is of particular interest (as discussed an arbitrary encoding function can associate any knowledge base to any network, but this will have no relevance to the ability of the network to learn a dataset).

Neuro-symbolic AI has potential to overcome many of the limitations of a purely data-driven approach to AI. As the practical work in this respect progresses and expands, a robust theory of the semantics of neuro-symbolic computation will be needed. With this paper we have aimed to provide the first steps in this direction.



%
%

\bibliographystyle{plainnat}
\bibliography{biblio}   
\appendix
\section{Proof of properties from Section 3.4.}
Proof of Theorem \ref{theorem:trans_1}:
\begin{proof} Take any $x_0\in X$, let $y_0=f(x_0)$. Because $N_2$ is a semantic encoding of $L$ under $i$, there exists $t_0$ such that $\forall t>t_0$, $N_2^t(y_0)=N_2^t(f(x_0))$ is a model of $L$. By assumption and induction on $t$ we have $N_2^t(f(x_0))=f(N_1^t(x_0))$ so $N_1^t(x_0)$ is a model of $L$ under $i\circ f$. Therefore, for all $x_0$, there exists $t_0$ such that $\forall t>t_0$, $N_1^t(x_0)$ is a model for $L$ under $i\circ f\in I$ and thus $N_1$ is a model for $L$ under $i\circ f \in I$. 
Now we must show that $L\vDash_{N_1} L' \implies L \vDash_{\mathcal{S}} L'$. Because $N_2$ is a semantic encoding of $L$, we have by definition $L\vDash_{N_2} L' \implies L \vDash_{\mathcal{S}} L'$, so we will show $L\vDash_{N_1} L' \implies L \vDash_{N_2} L'$ and that will complete the proof.
Let $X_m$ be the set of $L$-models of $N_1$ and $Y_m$ be the set of $L$-models of $N_2$. $L\vDash_{N_1} L'$ means that $L'$ is true in every $L$-model of $N_1$, in other words, every element of $i\circ f(X_m)$ is a model of $L'$. If we can show that $Y_m=f(X_m)$ then the $L$-models of $N_2$ map to $i\circ f(X_m)$ which we just showed consists entirely of models for $L'$ and thus we have $L \vDash_{N_2} L'$. Thus, to complete the proof we show that $Y_m=f(X_m)$.
  Let $x$ be an $L$-model of $N_1$ and $y=f(x)$. This means that $i\circ f(x)=i(y)$ is a model of $L$ and thus $y$ is a model of $L$ under $i$. Because $x$ is an $L$-model, there exists $x_0$ such that for all $t>0$, $\exists t'>t$ with $N_1^{t'}(x_0)=x$, this gives us $f(N_1^{t'}(x_0))=f(x)$ and by assumption $f(N_1^{t'}(x_0))=N_2^{t'}(f(x_0))$. Thus for all $t>0$, $\exists t'>t$ and $y_0=f(x_0) \in Y$ such that $y=N_2^{t'}(y_0)$ and thus $y$ is an $L$-model of $N_2$. 
Now let $y$ be an $L$-model of $N_2$, because $f$ is bijective, we can repeat the same argument above to show that $x_0=f^{-1}(y_0)$ is an $L$-model of $N_1$ and thus $Y_m=f(X_m)$ which completes the proof
\end{proof}
Proof of Theorem \ref{theorem:trans_2}:
\begin{proof}
    Let $L$ be a knowledge-base in $\mathcal{S}_2$ with models $\mathcal{M}_L$. Suppose $N$ encodes $f(L)$ under $i$, then $\mathcal{M}_N$ are models of $f(L)$. Because $g(g^{-1}(\mathcal{M}_N$)=$\mathcal{M}_N$ are models of $f(L)$, by assumption, $g^{-1}(\mathcal{M}_N)$ are models of $L$ and thus $N$ is a neural model of $L$ under $g^{-1}\circ i$. Now take another knowledge-base $L'\in \mathcal{L}_2$ and suppose $g^{-1}(\mathcal{M}_N)\subseteq \mathcal{M}_{L'}$, then by the same assumption as previously $\mathcal{M}_N\subseteq \mathcal{M}_{f(L')}$, because $N$ is a semantic encoding of $L$ this implies that $f(L)\vDash_{\mathcal{S}_1} f(L')$ which, by assumption, implies $L \vDash_{\mathcal{S}_2}L'$ meaning $N$ is a semantic encoding of $L$ under $g^{-1}\circ i$
\end{proof}
The proof of Corolory 3.5 is immediate.
\section{Proof that KBANN is semantically equivalent to Horn clauses}
\begin{proof}
KBANN provides the neural encoding method which we now outline. Given a set of acyclic Horn clauses, $P$, translate $P$ into a feed-forward network, $N$, with an encoding in $I_{NAT}$ by adding a neuron to the network for each atom $A_i$ that appears in $P$, mapping a state of the network $x$ to the interpretation $\{A_i | x_{A_i}=1\}$. Each Horn clause is encoded into the network by connecting each neuron representing atoms in the body to a neuron representing the atom in the head, setting the weights to implement a logical AND-gate. If an atom is in the head of multiple clauses then introduce a hidden neuron for each clause and connect the neurons representing the atoms in the body to that hidden neuron, and that hidden neuron to the neuron representing the atom in the head, setting the weights to implement a logical OR-gate.\footnote{This is because it can be proved that, without adding a hidden neuron, it is impossible to distinguish the valid combinations of the atoms in the bodies of the clauses. For example, given $A \leftarrow (B \wedge C)$ and $A \leftarrow (D \wedge E)$, it is impossible to connect neurons labelled as $B,C,D,E$ directly to a neuron labelled as $A$ such that firing $B$ and $C$ will fire $A$, firing $D$ and $E$ will fire $A$, but firing $B$ and $D$ or any other combination of $B,C,D,E$ does not fire $A$. Therefore, in this case, neurons $B$ and $C$ are first connected to a hidden neuron that is connected to $A$, and neurons $D$ and $E$ are connected to another hidden neuron that is connected to $A$.} If an atom is in the head of a clause with an empty body (i.e. it is a \emph{fact}) then it is given a bias that will always activate the corresponding neuron with no connections (i.e. zero-weight connections) from other neurons. We know that $T_P$ always converges to a unique fixed point for acyclic Horn clauses. The unique model of $P$ is the single fixed point of $T_P$ which is encoded by the network in state $\hat{x}$, where $\hat{x}_A=1$ if and only if there exists a clause $A\leftarrow B_1\wedge B_2 \wedge ... \wedge B_k$ with $\hat{x}_{B_i}=1$ for $i\in \{1,..,k\}$, $\hat{x}_A$ being the value of the neuron corresponding to atom $A$, and $\hat{x}_{B_i}$ the value of the neuron corresponding to atom $B_i$. Because $P$ is acyclic, $N$ is feed-forward and has a unique fixed point, which is $\hat{x}$, and thus $N$ is a semantic encoding of $P$ under $I_{NAT}$ and $Agg=\cup$.

We have seen that every set of acyclic Horn clauses can be semantically encoded into a feed-forward neural network with positive weights under $I_{NAT}$. Now we prove the converse. Take a feed-forward network, $N$, with positive weights and a state space $X=\{0,1\}^n$. We will construct a set of acyclic Horn clauses, $P$, that encodes $N$ under $I_{NAT}$ and $Agg=\cup$. First, we define an encoding function in $I_{NAT}$ by associating a propositional atom, $X_j$, to each neuron $j$ using the same encoding, $i(x)=\{X_j | x_j=1\}$. Given a state $x\in X$, for each neuron, $l$, if $x_l=1$ in $N(x)$ then we add the following clause to $P$: $X_l\leftarrow X_{j_1}\wedge X_{j_2} \wedge ... \wedge X_{j_k}$, where neurons $j_1,j_2,...,j_k$ are the input neurons to $l$ with $x_{j_i}=1$ in $x$. We repeat this process for each state $x\in X$. Because all the weights are positive, if $N(x)_l=1$ for a state $x$ then $N(x')_l=1$ if $x_i=1$ implies that $x_i'=1$. This guarantees the consistency of $P$ as it means that \textit{every} state that satisfies the body of the clause added to $P$ will activate neuron $l$ in the next time step, meaning that every clause added is valid over the entire state space. 

We can see that $P$ is a set of acyclic Horn clauses because $N$ is feed-forward. This means that $T_P$ always converges to a unique fixed point and clearly $\mathcal{M}_P\subseteq range(i)$, so in order to satisfy the conditions in Theorem \ref{lemma:tp}, all that is left to prove is $T_P(i(x))=i(N(x))$. Take any state, $x$, and consider $T_P(i(x))=\{X_l|\exists(X_l\leftarrow X_{j_1}\wedge ... \wedge X_{j_k})\in P, X_{j_i}\in i(x)\}$. Because $X_{j_i}\in i(x)$, $x_{j_i}=1$, and by definition, if $X_l\leftarrow X_{j_1}\wedge ... \wedge X_{j_k} \in P$ then $x_{j_1},x_{j_2},...,x_{j_k}=1$ implies that $N(x)_l=1$, which implies that $X_l\in i(N(x))$. Hence, if $X_l\in T_P(i(x))$ then $X_l\in i(N(x))$. Conversely, if $X_l\in i(N(x))$ then $N(x)_l=1$ and by definition there is a clause in $P$, $X_l\leftarrow X_{j_1}\wedge ... \wedge X_{j_k})$ with $x_{j_i}=1$, meaning that $X_{j_i}\in i(x)$ and thus $X_l\in T_P(i(x))$. Therefore, $T_P(i(x))=i(N(x))$.
\end{proof}
\section{Proof that LTN is an approximate encoding under $I_{DAT}$, $\cap$, and fuzzy logic fidelity.}
\begin{proof} First we show how an LTN is constructed. Given a Real Logic knowledge-base that we wish to model, $L$, an LTN is constructed by adding $n$ visible units to the network for each variable that appears in L. The neurons representing variables are considered input neurons and, given an initial state, stay fixed over time. 
We add $n$ neurons representing each complex term, $\theta$, that appears in $L$ recursively in the following way: if $\theta$ is of the form $f(\theta_1,...,\theta_k)$ then we add $n$ neurons for $\theta$ with connections to the $n \times k$ neurons representing $\theta_1,...,\theta_k$. These weights are determined solely by $f$. If $\theta'=f(\theta_1',...,\theta_k')$ is another term in $L$ then the weights connecting $\theta'$ to $\theta_1',...,\theta_k'$ are identical to those connecting $\theta$ to $\theta_1,...,\theta_k$. Each constant term $c$ is assigned a fixed vector $\bar{c}$.  Finally, we add a single visible unit for each predicate and argument combination appearing in the knowledge-base, $R(\theta_1,\theta_2,...,\theta_k)$, where $\theta_k$ is the term that is the $k^{th}$ argument for this instance of $R$. Each of these neurons is connected to the neurons representing its input terms. As with the function symbols, the weights are completely determined by $R$. Let $\hat{x}$ be the set of variables in $L$, and $\hat{c}$ be the set of constants.

The visible units are the neurons representing variables and predicates. The hidden neurons are all the other terms, i.e. those built by variables, functions and constants. Define our encoding function, $i$, by mapping each state to the set of groundings, $G$, which satisfy the following:
\begin{itemize}
    \item Let $x_{R\theta_1,...,\theta_k}$ be the value of the neuron representing $R(\theta_1,\theta_2,...,\theta_k)$;
    \item Let $\bar{x}_j$ be the vector of $n$ neurons representing $\hat{x}_j$;
    \item A state of $N$ maps to the set of groundings, $G$, that satisfy the following: $G(c)=\bar{c}$ for all constants $c$ that appear in $L$, 
    $G(f)$ is the function defined by the weights associated with $f$, and if $Q$ is an atom of the form $R(\theta_1(\bar{x},\bar{c}),...,\theta_k(\bar{x},\bar{c}))$ with $R(\theta_1,...\theta_k)\in L$, then $G(R(\theta_1(\bar{x},\bar{c}),...,\theta_k(\bar{x},\bar{c})))=x_{R\theta_1,...,\theta_k}$,
\end{itemize}
where $R(\theta_1(\bar{x},\bar{c}),...,\theta_k(\bar{x},\bar{c}))$ is the ground atom obtained by substituting the free variables in $\theta$ with $\bar{x}$, assigning the constants the value $\bar{c}$, and computing the value of $\theta$ with $G(f)$.\\\\
First, we show that $i\in I_{DAT}$. The interpretations we consider are groundings, $G$, for which the network parameters define $G(f)$ and $G(c)$ for each function and constant symbol. Because each grounding is fully defined by its interpretation of function symbols, constant symbols and predicates, these groundings are in one-to-one correspondence with the set of truth assignments to the ground atoms of the form $R(t_1,...,t_k)$ where $t_i\in O$. We choose a subset of atoms, $A'$, to consist of all grounded atoms, $R(t_1,...,t_k)$, for which $R(\theta_1,...,\theta_k)$ appears in $L$ and $t_i=\theta_i(\bar{x},\bar{c})$ for some variable assignment $\bar{x}$ and the network's constant assignment $\bar{c}$. For simplicity, we assume that this set contains no duplicates. In other words, each atom has at most one representation of this form\footnote{If there are multiple ways to represent an atom, we repeat the above process by defining a new set of maps $o_2,h_2,r_2$ in the same way but using the duplicate representations to compute their value instead. Repeat this until there are no additional representations for the atom.}.
We explicitly define the mappings, $o,,gh,r$, required in Definition $\ref{def:IDAT}$ to prove that $i\in I_{DAT}$: $h$ and $o$ specify, for an atom in $A'$, which variable assignment will represent it, in particular, $o$ maps an atom $R(\theta_1(\bar{x},\bar{c}),...,\theta_k(\bar{x},\bar{c}))$ onto the neurons corresponding to the variables that appear in $\theta_1,...,theta_k$, $h$ maps an atom $R(\theta_1(\bar{x},\bar{c}),...,\theta_k(\bar{x},\bar{c}))$ to the vector consisting of the values of each variable, $\bar{x}_i$ that appear in $\theta_1,...\theta_k$ (so if there are $j$ variables each of which has a value in $\mathbb{R}^n$ then $h$ maps to the $j\cdot n$ dimensional vector of their values); $r$ maps $R(\theta_1(\bar{x},\bar{c}),...,\theta_k(\bar{x},\bar{c}))$ to the label of neuron $x_{R\theta_1,...,\theta_k}$; $g$ is the identity function on $\mathbb{R}$. Surjectivity of $r$ is immediate from the definition and the assumption of no duplicates. Furthermore, if $Q,Q'\in A'$ with $Q\neq Q'$ then $R(\theta_1(\hat{x}=t),...,\theta_k(\hat{x}=t))\neq R'(\theta'_1(\hat{x}=t'),...,\theta'_k(\hat{x}=t'))$. If $t\neq t'$ then $h(Q)\neq h(Q')$; otherwise, if $R \neq R'$ or $\theta_i'\neq \theta_i$ then $r(Q)\neq r(Q')$. This proves the second condition. The final condition is satisfied by definition of $i$, in particular, if a grounding is in $i(x)$ then for an atom $Q=R(\theta_1(\hat{x}=\bar{x},\bar{c}),...,\theta_k(\hat{x}=\bar{x},\bar{c}))$, if $h(Q)=x_{o(Q)_1,...,o(Q)_m}$ then $x_j=\bar{x}_j$ for each variable $\hat{x}_j$ that appears in $\theta_1,...,\theta_k$ and by definition $G(R(\theta_1(\hat{x}=\bar{x},\bar{c}),...,\theta_k(\hat{x}=\bar{x},\bar{c})))=x_{R\theta_1,...,\theta_k}$, but $x_{R\theta_1,...,\theta_k}=g(x_{r(R(\theta_1(\hat{x}=\bar{x},\bar{c}),...,\theta_k(\hat{x}=\bar{x},\bar{c})))})$ so the conditions for $I_{DAT}$ are satisfied.\\\\
With our encoding in $I_{DAT}$, we now must look at $X_{inf}$. Define the computation time $t_c$ to be the integer corresponding to the maximum depth of the graph associated with the network. Because the LTN is feedforward, it will converge to a fixed point in which $\bar{x}$ is equal to its initial state, each hidden unit representing a term $\theta$ will be updated to $\theta(\bar{x},\bar{c})$ and $x_{R\theta_1,...,\theta_k}$ will be equal to the truth value obtained by computing $R(\theta_1(\bar{x}=t,\bar{c}),...,\theta_k(\bar{x}=t,\bar{c}))$ using the network parameters for $R$. With $Agg=\cap$, $\mathcal{M}_N$ represents the partial grounding in which every atom that can be represented by $R(\theta_1(\bar{x},\bar{c}),...,\theta_k(\bar{x},\bar{c}))$ for some $\bar{x}$ and $R(\theta_1,...,\theta_k)\in L$ has truth value equal to $x_{R\theta_1,...,\theta_k}$ in the stable state with input variables $\bar{x}$.  

This shows that an LTN defines a partial grounding, $\hat{G}$, on a knowledge-base, $L$. The loss function used to train the LTN is given in \cite{LTNs} as $\SatAgg\limits_{\phi \in L} \hat{G}(\phi)$ where $\hat{G}(\phi)$ is the fuzzy truth value assigned to $\phi$ by $\hat{G}$ using the chosen fuzzy connectives. This is equivalent to $\SatAgg\limits_{\phi \in L} 1-1- \hat{G}(\phi)=\SatAgg\limits_{\phi \in L} 1-d(\hat{G}(\phi),[1,1])$. Recalling that $\hat{G}$ represents a set of interpretations, each of which has the same value of $M(\phi)$, we can write $\hat{G}$ as $\mathcal{M}_N$  giving us $\inf\limits_{M \in \mathcal{M}_N} \SatAgg\limits_{\phi \in L} 1-d(M(\phi),[1,1])$, which is just $Fid_{fuzzy}((N,i,\cap),L)$. 
\end{proof}

\section{Proof that Semantic Loss is an approximate encoding under $I_{NAT}$, $\cup$, and probabilistic fidelity.} 
\begin{proof}
First we describe the networks under consideration and their encoding functions. Assume that we have a deep feed-forward neural network with $k$ layers whose first $k-1$ layers are deterministic and whose final layer consists of $n$ binary-valued neurons whose state is updated probabilistically according to a vector of probabilities $p=[p_1,...,p_n]$ computed from the state of the previous layer. Assume as before that the input layer is fixed for a given initial state, $x_0$. Denote by $p(x_0)$ the vector of probabilities of the output layer after $k-1$ time steps. Let the output neurons be visible and the rest hidden. Define $i\in I_{NAT}$ by $i(x)=\{M|M(Q_j)=True$ if $x_j=1$ and $M(Q_j)=False$ if $x_j=0\}$ where $Q_j$ is a propositional atom corresponding to the $j^{th}$ output neuron. $Loss(L,p(x_0))$ is simply the negative log probability that $i(X^{(k)})$ is a model of $L$. Minimizing this for all input states $x_0$ is thus equivalent to maximizing the probability that the state of the output neurons after $k-1$ time steps represents is a model of $L$ given any input $x_0$. Given a uniform initial distribution, this is equivalent to maximizing the probability that $x\in X_{P,inf}$ is a model of $L$ which is exactly the definition of  $Fid_{prob}((N,i,\cup),L)$.
\end{proof}

\end{document}